# MSK-LLaVA: A Multi-Image Vision–Language Model for Fracture Detection and Report Generation in Musculoskeletal Radiography

Seowoo Lee, MD[1]; Jihun Kim, MS[2,3]; Jee Won Chai, MD, PhD[4,5]; Hyo Jin Kim, MD, PhD[4,5]; Jiwoon Seo, MD, PhD[4,5]; Yoon-Hee Choi, MD, PhD[6]; Hyun Beom Song, MD, PhD[7]; Sung Hwan Hong, MD, PhD[1]; Young-Gon Kim, PhD[8,9]†; Dong Hyun Kim, MD[4,5]†

[1] Department of Radiology, Seoul National University College of Medicine, Seoul National University Hospital, Seoul, Republic of Korea.

[2] Department of Transdisciplinary Medicine, Seoul National University Hospital, Seoul, Republic of Korea.

[3] Department of Applied Bioengineering, Graduate School of Convergence Science and Technology, Seoul National University, Suwon, Republic of Korea.

[4] Department of Radiology, SMG-SNU Boramae Medical Center, Seoul, Republic of Korea.

[5] College of Medicine, Seoul National University, Seoul, Republic of Korea.

[6] Department of Physical Medicine and Rehabilitation, Soonchunhyang University Seoul Hospital, Soonchunhyang University College of Medicine, Seoul, Republic of Korea.

[7] Department of Biomedical Sciences, Seoul National University College of Medicine, Seoul, Republic of Korea.

[8] Department of Transdisciplinary Medicine, Seoul National University Hospital, Department of Medicine, Seoul National University College of Medicine, Seoul, Republic of Korea.

[9] Healthcare AI Research Institute, Seoul National University Hospital, Seoul, Republic of Korea.

**† Co-corresponding authors**

## Correspondence

**Dong Hyun Kim, MD**

Department of Radiology

SMG-SNU Boramae Medical Center

20 Boramae-ro 5-gil, Dongjak-gu

Seoul 07061, Republic of Korea

Tel: +82-2-870-3534

Email: mi4ri4@snu.ac.kr

**Young-Gon Kim, PhD**

Department of Transdisciplinary Medicine

Seoul National University Hospital

101 Daehak-ro, Jongno-gu

Seoul 03080, Republic of Korea

Tel: +82-2-2072-4487

Email: younggon2.kim@gmail.com

**Type of Manuscript:** Original Research

**Word Count for Text:** 2314

**Funding:** This study was supported by the NAVER Digital Bio Innovation Research Fund, funded by NAVER Corporation (Grant No. 3720230240)

**Conflicts of Interest:** The authors declare no competing interests. The funder had no role in the study design, data collection, analysis, interpretation, or the writing of the manuscript.

# MSK-LLaVA: A Multi-Image Vision-Language Model for Fracture Detection and Report Generation in Musculoskeletal Radiography

**Article type:** Original Research

## Summary Statement

**A multi-image musculoskeletal vision-language model achieved higher report-derived fracture detection performance than comparator models and showed potential to support physician interpretation of musculoskeletal radiographs.**

## Key Points

A multi-image vision-language model developed for musculoskeletal radiography (MSK-LLaVA) achieved report-derived fracture F1 scores of 0.86 and 0.93 in the internal and external test sets, respectively, compared with 0.24–0.48 and 0.08–0.43 for seven general-purpose and medical VLMs.

In a crossover reader study involving three nonradiologist physicians, AI-assisted interpretation increased fracture detection accuracy from 88.2% to 94.3% ($P = .016$) and sensitivity from 76.6% to 94.3% (nominal $P < .001$), with 16% shorter adjusted interpretation time (nominal $P = .004$).

In the internal test set, 57% of MSK-LLaVA reports were acceptable without revision, 24% required minor revision, and 19% required major revision or were unacceptable.

## Abstract

### Purpose

To develop and evaluate MSK-LLaVA, a multi-image musculoskeletal vision-language model for fracture detection and report generation, and assess AI-assisted fracture detection among nonradiologist physicians.

**Materials and Methods**

This retrospective study included 1,404,105 radiographs from 141,241 patients during 2010–2023 (mean age, 59.7 years; 76,306 female patients), with temporal partitioning. Anatomically refined reports provided radiograph-specific supervision. MSK-LLaVA combined the musculoskeletal encoder MSK-SigLIP with a 1.7-billion-parameter language model to process up to eight radiographs per examination. Models were compared on 100 internal and 116 external examinations. Three nonradiologist physicians completed a crossover study. Reader analyses used examination-level paired differences and case-cluster-robust regression.

**Results**

MSK-SigLIP achieved the highest linear-probe fracture area under the receiver operating characteristic curve among evaluated encoders internally (0.90), on FracAtlas (0.98), and on GRAZPEDWRI-DX (0.97). It also had the highest internal estimates for arthritis (0.84), osteopenia (0.86), and soft-tissue swelling (0.94). MSK-LLaVA achieved report-derived fracture F1 scores of 0.86 (95% confidence interval [CI]: 0.76, 0.94) internally and 0.93 (95% CI: 0.88, 0.98) externally, exceeding comparator ranges of 0.24–0.48 and 0.08–0.43. Assistance increased fracture accuracy from 88.2% to 94.3% ($P = .016$) and sensitivity from 76.6% to 94.3% (nominal $P < .001$). Specificity was 96.1% and 94.2%, respectively (nominal $P = .35$). Adjusted interpretation time was 16% shorter (ratio, 0.84; 95% CI: 0.75, 0.94; nominal $P = .004$).

**Conclusion**

MSK-LLaVA achieved higher report-derived fracture detection performance than general-purpose and medical comparators. AI-assisted interpretation may improve fracture detection among nonradiologist physicians.

## Introduction

Plain radiography is a first-line examination for suspected fractures in emergency and outpatient care. Interpretation requires more than identifying a fracture: radiologists must determine its location, displacement, and articular involvement while assessing alignment, soft tissues, degenerative changes, and surgical hardware. Subtle cortical abnormalities and overlapping anatomy can make this difficult, particularly when timely specialist interpretation is unavailable (1, 2).

Deep learning has expanded automated fracture detection, while vision-language models (VLMs) offer the possibility of generating free-text reports (3, 4). Musculoskeletal (MSK) reporting nevertheless poses specific challenges. Abnormalities may be small relative to the image, and their interpretation often depends on complementary projections. Encoders trained on natural images or heterogeneous medical images may not adequately represent these features (5). Models must also assign findings to the correct anatomic region and side when integrating several views.

Radiologic reports may describe multiple anatomic regions or both sides, whereas an individual radiograph covers only part of the reported anatomy. Pairing each radiograph with the full report can therefore associate it with findings outside its field of view or on the opposite side. Refining reports according to expected anatomic coverage and laterality may reduce this mismatch and provide more appropriate radiograph-specific supervision.

The purpose of this study was to develop and evaluate MSK-LLaVA, a multi-image VLM for fracture detection and examination-level report generation in musculoskeletal radiography (Supplemental Methods 1). We compared its report-derived detection performance with that of general-purpose and medical VLMs and assessed the quality of its generated reports. In a reader study, we assessed whether AI support that included MSK-LLaVA reports improved fracture detection and reduced interpretation time among nonradiologist physicians. Separately, we evaluated MSK-SigLIP, the musculoskeletal image encoder developed for MSK-LLaVA, on internal and external datasets.

## Materials and Methods

### Study Design and Data Sources

This retrospective study received institutional review board approval with a waiver of written informed consent. Single-institution radiographs from 2010–2020 and 2021–2022 formed training and validation sets; internal testing used 2023 examinations from patients absent from both sets (Figure 1; Supplemental Tables 1–2, Supplemental Methods 2).

For final fine-tuning, four MSK radiologists prepared reports and labels for 3000 emergency department examinations performed before 2023. Internal testing included 100 randomly selected emergency department examinations covering one anatomic region, including contralateral views when available. Two MSK radiologists established six binary examination-level reference standards and radiograph-level labels for a random subset of 200 of the 535 internal test radiographs by consensus using radiographs and available follow-up imaging. The external test set comprised 116 examinations randomly selected from musculoskeletal radiographic examinations performed in the emergency department of an independent tertiary hospital in 2023; examinations acquired with protocols not included in this study (Supplemental Table 2) were excluded. Two MSK radiologists established reference binary labels by consensus using radiographs and radiologic reports; CT, MRI, and clinical records were unavailable.

### Report Refinement and Model Training

Llama 3.3 (70B) standardized radiologic reports and generated radiograph-specific text using protocol, laterality, and expected anatomic coverage, without image inputs (6). Prompts instructed removal of out-of-field, contralateral, and radiographically occult findings; eligible CT or MRI reports supplied auxiliary context. Unmentioned findings were coded as negative for training (Supplemental Methods 3).

MSK-SigLIP underwent supervised fracture classification followed by SigLIP-style alignment with radiograph-level reports (7). MSK-LLaVA linked this encoder to Qwen3 1.7B through a multilayer perceptron projector (8, 9). Training comprised projector-only alignment, full-model fine-tuning with single-radiograph and multi-radiograph inputs, and final fine-tuning on expert-annotated data. Inputs included protocol and laterality information (Figure 2, Supplemental Methods 4). Examinations exceeding eight radiographs used eight radiographs randomly sampled during training and the first eight in fixed order during inference.

## Evaluation of MSK-SigLIP and Comparator Encoders

Encoder evaluation used the internal 200-radiograph subset and five external public test sets: MURA, FracAtlas, GRAZPEDWRI-DX, LERA, and BTXRD (10-14). Zero-shot classification used predefined prompts with paired text encoders. Linear probing fitted logistic regression to normalized features from frozen encoders. Dataset splits, cross-validation, metrics, and aggregation are detailed in Supplemental Methods 5 and Supplemental Tables 3–4 (5, 15-19).

### Evaluation of MSK-LLaVA and Comparator Report-Generation Models

Two general-purpose and five medical VLMs received identical radiographs and semantically equivalent instructions without fine-tuning (15, 20-25). Model-specific inference settings are detailed in Supplemental Methods 6. All comparator models were open-weight models run on local hardware. Proprietary models available only through external application programming interfaces (APIs), such as ChatGPT and Gemini, were not evaluated for image interpretation because institutional policy prohibited transferring patient radiographs outside the hospital network.

GPT-4o mini extracted six findings: fracture, soft-tissue swelling, arthritis, prosthetic hardware, osteopenia, and joint dislocation. Because only model-generated text containing no protected health information was submitted, external API processing was permitted for structured label extraction. Uniform rules classified possible or suspected findings as positive and unmentioned findings as negative. Human validation of the structured labeler was performed using 100 internal MSK-LLaVA reports; automated labels were retained without correction (Supplemental Methods 7). Sensitivity, specificity, and F1 scores were calculated against radiologist-established references. Two MSK radiologists independently rated these reports as acceptable without revision, requiring minor revision, requiring major revision, or unacceptable, resolving disagreements by consensus.

## Reader Study

Three nonradiologist physicians with less than 1 year of emergency department experience interpreted all 116 external examinations with and without assistance. Examinations were randomized 1:1 to assisted-first or unassisted-first interpretation, with the same allocation used for all readers. Sessions were separated by 3 weeks; the order of examinations was randomized independently for each session, and readers could not access their previous responses. Assistance added MSK-LLaVA reports and gradient-weighted class activation maps (Grad-CAM) from the fracture classifier to radiographs, age, and sex (26) (Supplemental Methods 8). For each examination, readers wrote a free-text report and recorded binary labels for the six findings (fracture, soft-tissue swelling, arthritis, prosthetic hardware, osteopenia, and joint dislocation). Interpretation time was measured from radiograph display to submission of the report and labels. Fracture accuracy was the primary outcome; sensitivity, specificity, and interpretation time were supporting outcomes.

## Statistical Analysis

Encoder performance was assessed using the area under the receiver operating characteristic curve (AUC) and F1 score. Report-derived finding detection was evaluated using sensitivity, specificity, and F1 score against radiologist-established references. Model-performance 95% confidence intervals (CIs) used 2000 bootstrap resamples. Labeler agreement and report-quality ratings were summarized descriptively.

Reader-study effects were expressed as assisted-minus-unassisted differences in fracture accuracy, sensitivity, and specificity, averaged across readers within examinations, with t-based 95% CIs and tests. Adjusted analyses used logistic generalized estimating equations for diagnostic performance and regression of log-transformed interpretation time, yielding odds ratios and geometric mean time ratios with case-cluster-robust inference. The primary accuracy test used two-sided $P < .05$; supporting analyses were exploratory, and nonfracture findings were descriptive (Supplemental Methods 9).

## Results

### Cohort Characteristics

The initial cohort comprised 220,295 patients, 2,408,423 examinations, and 3,568,988 radiographs. Exclusions removed 2,057,799 radiographs from nontarget protocols (Supplemental Table 2), 65,498 with integrity or metadata problems, and 41,586 without reports. The eligible source cohort contained 141,241 patients (mean age, 59.7 ± 20.9 years; 76,306 female patients [54.0%]), 928,839 examinations, and 1,404,105 radiographs before test-patient exclusions and sampling (Table 1). Detailed imaging characteristics are provided in Supplemental Table 1.

Training included 125,319 patients, 808,262 examinations, and 1,221,436 radiographs; validation included 13,322 patients, 64,964 examinations, and 99,879 radiographs. Internal testing comprised 100 patients, 100 examinations, and 535 radiographs. The external set contained 116 examinations and 213 radiographs, including 47 fracture-positive examinations (40.5%).

### Encoder Performance

MSK-SigLIP had the highest fracture AUC point estimates among evaluated encoders in zero-shot classification: 0.87 (95% CI: 0.77, 0.94) internally, 0.84 (95% CI: 0.82, 0.86) on FracAtlas, and 0.92 (95% CI: 0.915, 0.923) on GRAZPEDWRI-DX (Supplemental Table 5). Corresponding linear-probe AUCs were 0.90 (95% CI: 0.82, 0.96), 0.98 (95% CI: 0.96, 0.99), and 0.97 (95% CI: 0.96, 0.97). Beyond fracture detection, MSK-SigLIP also achieved the highest linear-probe AUC point estimates for arthritis (0.84), osteopenia (0.86), and soft-tissue swelling (0.94) in the internal test set, abnormality detection on MURA (0.91), and tumor detection on BTXRD (0.95) (Supplemental Table 6). However, other encoders had higher linear-probe AUC point estimates for joint dislocation (SigLIP 2, 0.72; MSK-SigLIP, 0.67) and prosthetic hardware (DINOv3, 0.99; MSK-SigLIP, 0.87) in the internal test set, abnormality detection on LERA (MedSigLIP, 0.86; MSK-SigLIP, 0.80), and malignancy classification in the BTXRD tumor-positive subset (SigLIP 2, 0.97; MSK-SigLIP, 0.95).

## Report Performance and Quality

GPT-4o mini–extracted labels agreed with the radiologists' consensus interpretation of the generated report text in 598 of 600 comparisons (99.7%) in the internal test set. The two discrepancies involved prosthetic hardware (Supplemental Methods 7). In the internal test set, MSK-LLaVA had the highest F1 point estimate for five of six findings (Table 2). Fracture sensitivity was 85.7% (30/35; 95% CI: 74%, 97%), specificity was 92.3% (60/65; 95% CI: 85%, 98%), and F1 score was 0.86 (95% CI: 0.76, 0.94). Comparator fracture F1 scores ranged from 0.24 (Qwen3.8) to 0.48 (Gemma 4). MedGemma (27B) had the highest prosthetic hardware F1 score (0.93). Sensitivity, specificity, and confusion-matrix counts for all models are provided in Supplemental Table 7.

In the external test set, MSK-LLaVA had the highest F1 point estimate for all six findings (Table 2). Fracture sensitivity was 95.7% (45/47; 95% CI: 89%, 100%), specificity was 92.8% (64/69; 95% CI: 86%, 99%), and F1 score was 0.93 (95% CI: 0.88, 0.98). Fracture F1 scores for the seven comparator models ranged from 0.08 to 0.43. Corresponding finding-level metrics and confusion-matrix counts for all models are provided in Supplemental Table 8.

Of the 100 MSK-LLaVA reports generated for the internal test set, 57 were acceptable without revision, 24 required minor revision, 8 required major revision, and 11 were unacceptable. On qualitative review, the radiologists identified 5 missed fractures, 4 false-positive fractures, 1 missed joint dislocation, and 1 fracture-laterality error among the unacceptable reports. Examples of correct findings, omissions, and unsupported findings are shown in Figures 3–6.

## Reader Performance

Assistance increased fracture accuracy from 88.2% (307/348) to 94.3% (328/348), a difference of 6.0 percentage points (95% CI: 1.1, 10.9; $P = .016$). Sensitivity increased from 76.6% (108/141) to 94.3% (133/141), a difference of 17.7 percentage points (95% CI: 8.0, 27.5; nominal $P < .001$). Specificity was 96.1% (199/207) without assistance and 94.2% (195/207) with assistance, a difference of −1.9 percentage points (95% CI: −6.0, 2.2; nominal $P = .35$). The adjusted population-averaged accuracy odds ratio was 2.19 (95% CI: 1.13, 4.23) (Table 3). Descriptive results for nonfracture findings and reader-specific fracture performance are provided in Supplemental Tables 9 and 10.

Median interpretation time was 38.0 seconds (interquartile range, 18.8–84.0) without assistance and 31.0 seconds (interquartile range, 15.0–62.0) with assistance. The adjusted geometric mean time ratio was 0.84 (95% CI: 0.75, 0.94; nominal P = .004) (Supplemental Table 11).

## Discussion

MSK-LLaVA integrated a musculoskeletal image encoder and anatomically refined supervision to generate reports from multiple radiographs. Its report-derived fracture F1 scores were higher than those of the evaluated general-purpose and medical VLMs in both test sets. Assistance combining generated reports and fracture activation maps was associated with higher fracture detection accuracy and sensitivity among nonradiologist physicians and shorter interpretation time. These findings suggest potential as a diagnostic aid, although clinically relevant report errors remained.

MSK-SigLIP showed strong classification performance across multiple musculoskeletal findings, including fractures, arthritis, osteopenia, soft-tissue swelling, and tumors. These results suggest that the encoder learned visual features relevant to a range of musculoskeletal abnormalities, providing a domain-specific foundation for MSK-LLaVA. Combining this encoder with a 1.7-billion-parameter language model supported multi-image report generation, although the contribution of individual model components requires further investigation.

Report refinement addressed the mismatch between individual radiographs and reports covering several regions or sides. Assigning text according to expected anatomy made supervision more specific, while subsequent multi-image training supported examination-level interpretation. However, expected anatomic coverage is a proxy for actual visibility. Because refinement did not inspect radiographs, it could not verify whether a described lesion was visible in a particular projection.

The reader findings are consistent with studies by Lindsey et al and Duron et al showing that fracture-detection assistance can improve clinician interpretation (1, 2). Among the three participating physicians, assistance was associated with a 6.0-percentage-point increase in fracture detection accuracy; supporting analyses suggested higher sensitivity and shorter interpretation time. Specificity decreased numerically, and the absence of statistical significance does not establish equivalence. Because assistance combined reports and activation maps, the benefit cannot be attributed to generated reports alone. Effects on false-positive decisions and subsequent management require prospective evaluation.

Report quality provided a more cautious picture than fracture F1 scores alone. Nineteen of 100 reports required major revision or were unacceptable, with errors including missed fractures, false-positive findings, and laterality errors. Binary finding extraction does not assess anatomic precision, morphology, or report completeness (27). Similarly, agreement between the automated labeler and radiologists supports the extraction procedure, not the clinical correctness of generated text. Finding-specific performance may also reflect differences in supervision: encoder pretraining explicitly emphasized fractures, whereas clinical reports may omit hardware or describe only reduced alignment in examinations containing both pre- and post-reduction radiographs. These are possible explanations for differences in performance across findings, but their contributions were not measured.

This study had several limitations. Development data came from one institution, and the test sets were small, particularly for uncommon findings; nonfracture outcomes were therefore reported descriptively. Excluded imaging categories restrict applicability. Report-derived negatives and auxiliary CT or MRI context may introduce supervision errors, while the external reference standard lacked cross-sectional or follow-up confirmation. Computational constraints precluded ablation studies to isolate the contributions of encoder training, refinement, multi-image input, and expert post-training. Differences in training data and preprocessing also limit architectural comparisons. Automated labeler validation was performed on internal MSK-LLaVA reports only. The reader study only included three physicians. Recall and learning effects could not be excluded despite the 3-week washout.

MSK-LLaVA achieved higher report-derived fracture F1 scores than the evaluated comparators.

Combined report and activation-map assistance may improve fracture detection among nonradiologist physicians, while errors in generated reports underscore the need for physician review.

## Code and Model Availability

The MSK-LLaVA model is available for download at https://github.com/ECOFRI/MSK-LLaVA.

## Acknowledgments

ChatGPT (OpenAI; accessed October 9, 2026) was used to assist with editing and organization of the manuscript. The authors reviewed and revised the text and take responsibility for its accuracy and content.

## References.

**Table 1. Demographic and Dataset Characteristics of the Eligible Internal Source Cohort**

| Characteristic | Value |
|---|---|
| **Patient characteristics** | |
| **No. of patients** | 141,241 |
| **Age, y, mean ± SD** | 59.7 ± 20.9 |
| **Male sex** | 64,935 (46.0) |
| **Female sex** | 76,306 (54.0) |
| **Dataset characteristics** | |
| **No. of examinations** | 928,839 |
| **No. of radiographs** | 1,404,105 |
| **Radiographs per examination, mean ± SD** | 1.51 ± 0.79 |
| **Radiographs per examination, range** | 1–16 |

Note. Values are counts unless otherwise indicated. Parentheses indicate percentages. The source cohort precedes patient-overlap exclusions and internal test sampling. SD = standard deviation.

**Table 2. F1 Scores for Report-Derived Findings in the Internal and External Test Sets**

| Model | Fracture | Soft-tissue swelling | Arthritis | Prosthetic hardware | Osteopenia | Joint dislocation |
|---|---|---|---|---|---|---|
| *Internal test set* | | | | | | |
| **MSK-LLaVA** | **0.86 (0.76, 0.94)** | **0.33 (0.09, 0.57)** | **0.58 (0.33, 0.80)** | 0.89 (0.76, 1.00) | **0.20 (0.00, 0.55)** | **0.67 (0.00, 1.00)** |
| **MedGemma 1.5 (4B)** | 0.34 (0.15, 0.51) | 0.00 (0.00, 0.00) | 0.00 (0.00, 0.00) | 0.80 (0.67, 0.94) | 0.00 (0.00, 0.00) | 0.00 (0.00, 0.00) |
| **MedGemma (27B)** | 0.30 (0.13, 0.46) | 0.00 (0.00, 0.00) | 0.12 (0.00, 0.35) | **0.93 (0.77, 1.00)** | 0.00 (0.00, 0.00) | 0.40 (0.00, 0.86) |
| **Lingshu (32B)** | 0.44 (0.28, 0.58) | 0.07 (0.00, 0.23) | 0.22 (0.09, 0.34) | 0.76 (0.62, 0.94) | 0.00 (0.00, 0.00) | 0.00 (0.00, 0.00) |
| **Fleming-VL (38B)** | 0.27 (0.10, 0.44) | 0.15 (0.00, 0.34) | 0.25 (0.06, 0.44) | 0.89 (0.76, 1.00) | 0.00 (0.00, 0.00) | 0.00 (0.00, 0.00) |
| **MedDr (40B)** | 0.27 (0.12, 0.43) | 0.00 (0.00, 0.00) | 0.00 (0.00, 0.00) | 0.67 (0.48, 0.86) | 0.00 (0.00, 0.00) | 0.00 (0.00, 0.00) |
| **Gemma 4 (31B)** | 0.48 (0.28, 0.64) | 0.07 (0.00, 0.21) | 0.14 (0.00, 0.38) | 0.80 (0.67, 0.94) | 0.00 (0.00, 0.00) | 0.17 (0.00, 0.35) |
| **Qwen3.8 (27B)** | 0.24 (0.06, 0.41) | 0.00 (0.00, 0.00) | 0.00 (0.00, 0.00) | 0.76 (0.62, 0.94) | 0.00 (0.00, 0.00) | 0.50 (0.00, 0.89) |
| *External test set* | | | | | | |
| **MSK-LLaVA** | **0.93 (0.88, 0.98)** | **0.50 (0.14, 0.77)** | **0.71 (0.52, 0.89)** | **0.93 (0.77, 1.00)** | **0.38 (0.27, 0.55)** | **0.50 (0.00, 1.00)** |
| **MedGemma 1.5 (4B)** | 0.21 (0.08, 0.36) | 0.00 (0.00, 0.00) | 0.00 (0.00, 0.00) | 0.76 (0.62, 0.94) | 0.00 (0.00, 0.00) | 0.00 (0.00, 0.00) |
| **MedGemma (27B)** | 0.26 (0.13, 0.39) | 0.00 (0.00, 0.00) | 0.00 (0.00, 0.00) | 0.40 (0.00, 0.77) | 0.00 (0.00, 0.00) | 0.00 (0.00, 0.00) |
| **Lingshu (32B)** | 0.43 (0.27, 0.57) | 0.10 (0.00, 0.29) | 0.18 (0.07, 0.30) | 0.62 (0.50, 0.76) | 0.00 (0.00, 0.00) | 0.00 (0.00, 0.00) |
| **Fleming-VL (38B)** | 0.08 (0.00, 0.19) | 0.00 (0.00, 0.00) | 0.33 (0.14, 0.51) | 0.76 (0.62, 0.94) | 0.00 (0.00, 0.00) | 0.00 (0.00, 0.00) |
| **MedDr (40B)** | 0.38 (0.21, 0.52) | 0.00 (0.00, 0.00) | 0.29 (0.00, 0.53) | 0.52 (0.31, 0.73) | 0.00 (0.00, 0.00) | 0.00 (0.00, 0.00) |
| **Gemma 4 (31B)** | 0.29 (0.12, 0.45) | 0.00 (0.00, 0.00) | 0.00 (0.00, 0.00) | 0.70 (0.55, 0.89) | 0.00 (0.00, 0.00) | 0.07 (0.00, 0.19) |
| **Qwen3.8 (27B)** | 0.19 (0.04, 0.32) | 0.00 (0.00, 0.00) | 0.12 (0.00, 0.35) | 0.73 (0.59, 0.89) | 0.00 (0.00, 0.00) | 0.00 (0.00, 0.00) |

Note. Data are F1 scores, with 95% confidence intervals in parentheses. Boldface indicates the highest unrounded point estimate for each finding among the models shown, determined separately for the internal and external test sets.

**Table 3. Effect of Assistance on Fracture Detection in the Reader Study**

| Metric | Unassisted | Assisted | Difference (percentage points) | P value | Adjusted OR |
|---|---|---|---|---|---|
| **Accuracy** | 307/348 (88.2%) (83.0, 92.8) | 328/348 (94.3%) (90.5, 97.4) | 6.0 (1.1, 10.9) | .016 | 2.19 (1.13, 4.23) |
| **Sensitivity** | 108/141 (76.6%) (66.0, 85.8) | 133/141 (94.3%) (88.7, 98.6) | 17.7 (8.0, 27.5) | <.001 | 5.29 (2.07, 13.54) |
| **Specificity** | 199/207 (96.1%) (92.8, 99.0) | 195/207 (94.2%) (89.4, 98.1) | -1.9 (-6.0, 2.2) | .35 | 0.43 (0.13, 1.46) |

Note. Reader values are n/N (%), followed by 95% confidence intervals (CIs) from 20,000 case-cluster bootstrap resamples. Differences are assisted minus unassisted, with 95% t intervals based on examination-level averages across the three readers. P values test these unadjusted differences using 116, 47, and 69 examinations for accuracy, sensitivity, and specificity, respectively. The accuracy P value is primary; sensitivity and specificity P values are nominal and exploratory. Adjusted population-averaged odds ratios (ORs) and 95% CIs are from logistic generalized estimating equations with assistance, period, sequence, and reader covariates and case-cluster-robust covariance; P values are not reported for the adjusted ORs. All inference is conditional on the participating readers and assumes independent examinations.

**Figure 1. Study cohort and evaluation overview.**

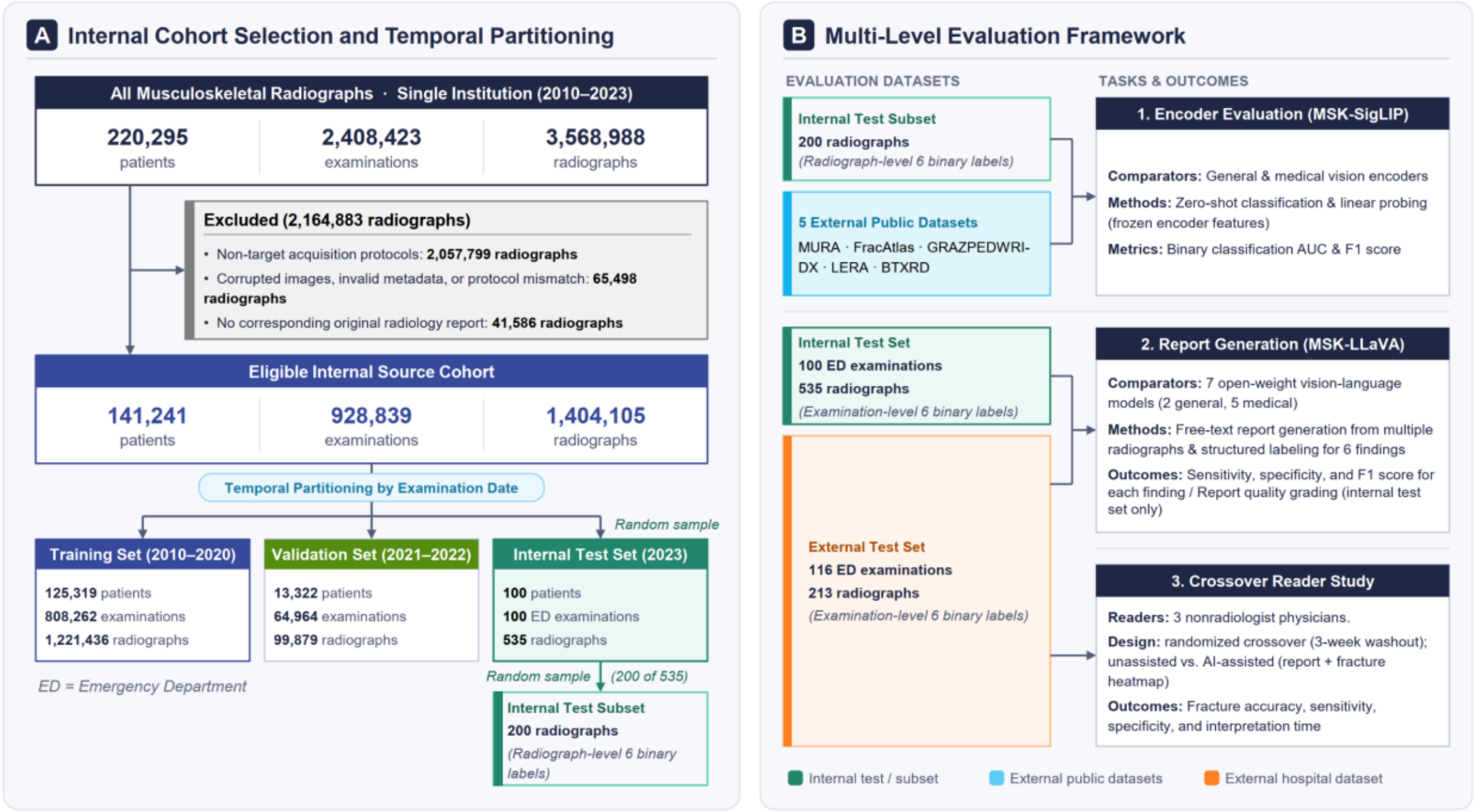

(A) Selection of the internal cohort and temporal partitioning. Training and validation data came from 2010–2020 and 2021–2022, respectively. The internal test set was 100 randomly sampled emergency department examinations from 2023, drawn from patients not included in training or validation. A random subset of 200 radiographs from this set received radiograph-level reference labels. (B) Datasets and outcomes for the three evaluations: image encoder evaluation, report generation, and the crossover reader study.

**Figure 2. Data refinement, model architecture, and training.**

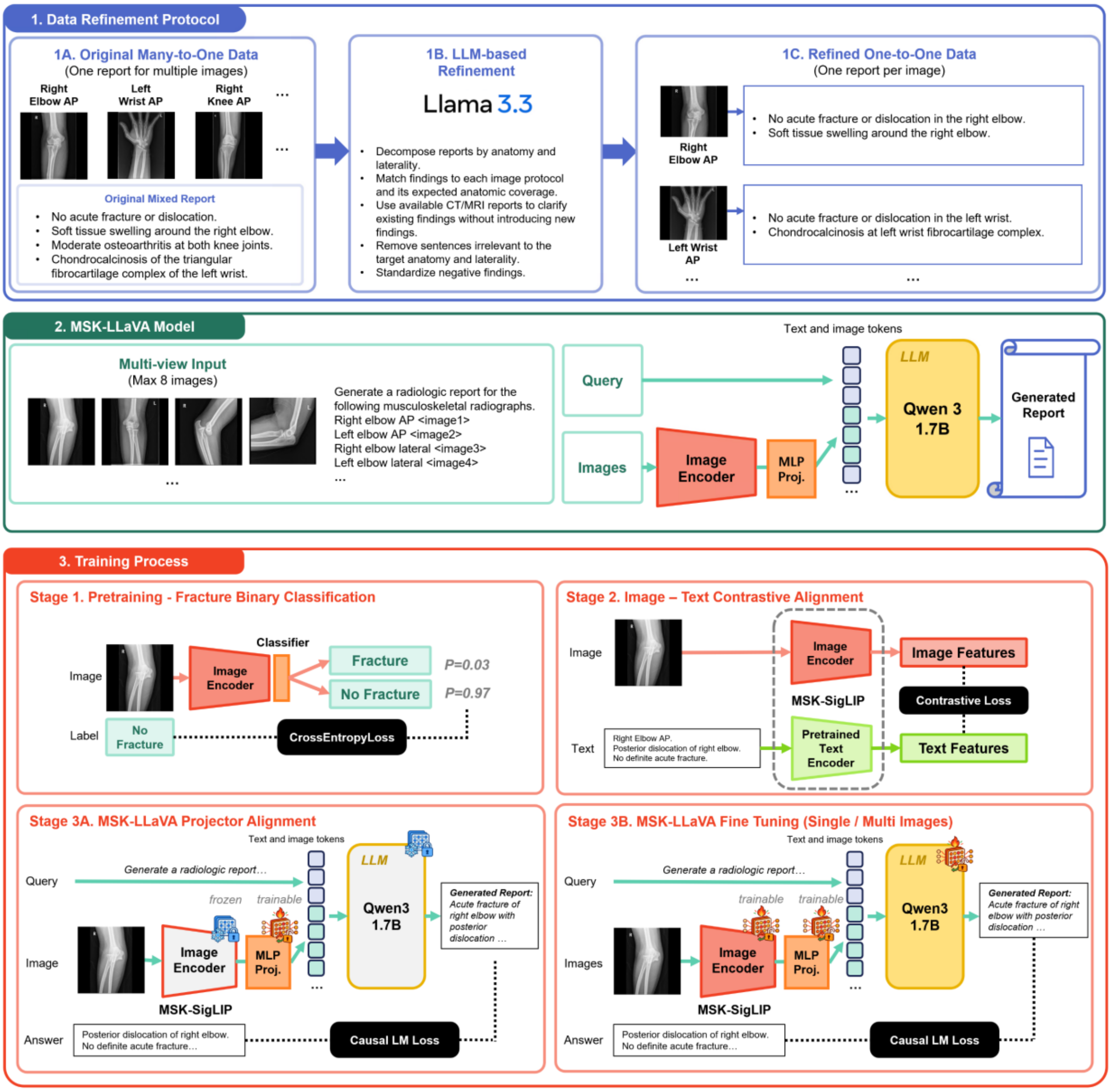

(1) Llama 3.3 (70B) refined each original report into radiograph-specific reports using protocol, laterality, and expected anatomic coverage, without access to radiographs. (2) MSK-LLaVA processes up to eight radiographs; MSK-SigLIP features are projected through an MLP and input to Qwen3 1.7B with text tokens. (3) Stage 1, fracture classification; Stage 2, image–text contrastive alignment, yielding MSK-SigLIP; Stage 3A, projector-only training; Stage 3B, full-model fine-tuning with single- and multi-radiograph inputs. Stage 3C (not shown) continued Stage 3B on expert-annotated data. Snowflake and flame icons indicate frozen and trainable components. P = predicted class probability, AP = anteroposterior, LLM = large language model, LM = language modeling, MLP =

multilayer perceptron, Proj. = projector.

**Figure 3. Fracture detection and report generation in the internal test set.**

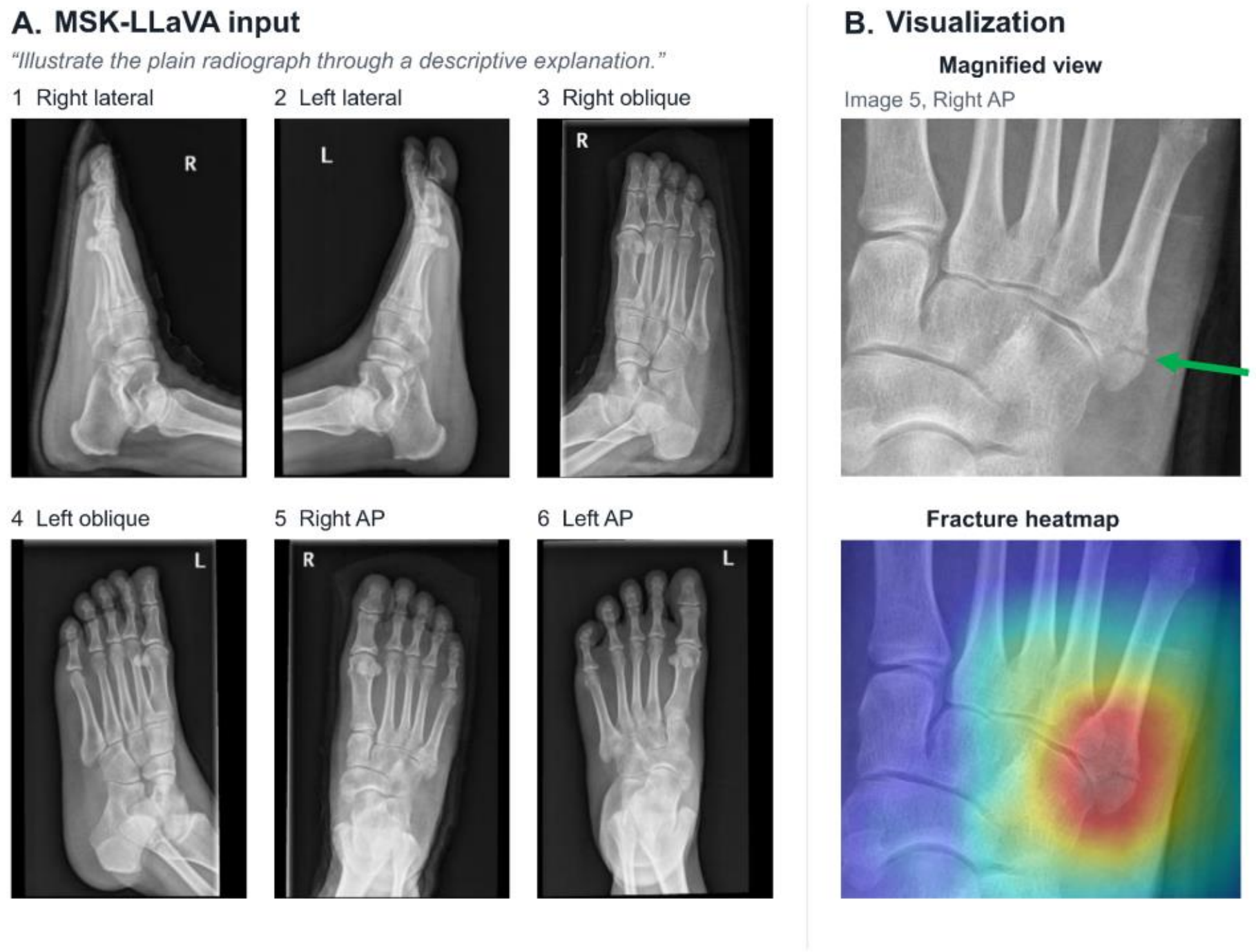

C. Report comparison

| | |
|---|---|
| **Reference standard** | **The right foot demonstrates a fracture at the base of the fifth metatarsal.** No soft tissue swelling, joint effusion, arthritis, osteoporosis or osteopenia, foreign body, prosthetic hardware, or joint dislocation is observed in the right foot. The left foot shows no fracture, soft tissue swelling, joint effusion, arthritis, osteoporosis or osteopenia, foreign body, prosthetic hardware, or joint dislocation. |
| **MSK-LLaVA**<br>Fracture reported | **There is a fracture at the base of the right fifth metatarsal bone.** No acute fracture is identified in the left foot. |
| **MedGemma 1.5 (4B)**<br>Fracture not reported | FINDINGS: The right foot demonstrates normal alignment. **There is no acute fracture or dislocation.** Joint spaces are preserved. No osseous lesions are identified. Soft tissues appear unremarkable. The left foot demonstrates normal alignment. There is no acute fracture or dislocation. Joint spaces are preserved. No osseous lesions are identified. Soft tissues appear unremarkable. IMPRESSION: No acute abnormality. |
| **MedGemma (27B)**<br>Fracture not reported | **No acute fracture, dislocation, or significant osseous abnormality is identified.** Joint spaces appear preserved. No significant soft tissue abnormality is seen. |
| **Lingshu (32B)**<br>Fracture not reported | Findings: The bones are of normal density. There is mild flattening of the arch consistent with pes planus. **No fractures, dislocations, or destructive lesions.**<br>There are small soft tissue calcifications adjacent to the right first metatarsal head which may represent plantar fibromatosis. This finding was not present on prior exam from 07/2019 but could be due to differences in technique between exams. Mild degenerative changes at the great toe IP joints bilaterally. |
| **Fleming-VL (38B)**<br>Fracture not reported | The radiographs of the right and left feet show **no acute fractures or dislocations**. The bones, including phalanges, metatarsals, tarsus, calcaneus, talus, navicular, cuboid, cuneiforms, and sesamoids are intact with normal alignment in both oblique and AP views. There is no evidence of joint space narrowing that would suggest osteoarthritis nor any signs of soft tissue swelling or abnormal calcifications. Both ankles demonstrate a normal articulation between the distal ends of the tibia and fibula with the talus bone without any visible abnormalities. |
| **MedDr (40B)**<br>Fracture not reported | **The bones are unremarkable.** |
| **Gemma 4 (31B)**<br>Fracture not reported | Radiographs of the right and left feet demonstrate **no acute fracture or dislocation**. Alignment is normal, and joint spaces are well-maintained. There are no osseous lesions or significant soft-tissue abnormalities. |
| **Qwen3.8 (27B)**<br>Fracture not reported | Multiple views of the right and left feet (AP, oblique, and lateral) demonstrate normal osseous alignment **without evidence of acute fracture or dislocation**. The metatarsals, phalanges, tarsal bones, and visualized distal tibia and fibula are intact with preserved cortical margins and no focal lytic or sclerotic lesions. The metatarsophalangeal, tarsometatarsal, midtarsal, subtalar, and ankle joint spaces are maintained without subluxation. No periosteal reaction, soft-tissue swelling, or abnormal calcification is identified. No acute radiographic abnormality is seen. |

Radiographs of a 52-year-old female patient show a fracture at the base of the right fifth metatarsal.

(A) Anteroposterior, oblique, and lateral radiographs of both feet used as model inputs. (B) Magnified right anteroposterior view showing the fracture (green arrow), with corresponding focal activation on the fracture heatmap. (C) Reference and model-generated reports. MSK-LLaVA correctly identified the fracture and specified its location and laterality. None of the seven comparator models reported the fracture. Blue text highlights descriptions of the right fifth metatarsal fracture. Orange text highlights statements negating fracture or describing the bones as unremarkable.

**Figure 4. Report generation for bilateral knee osteoarthritis.**

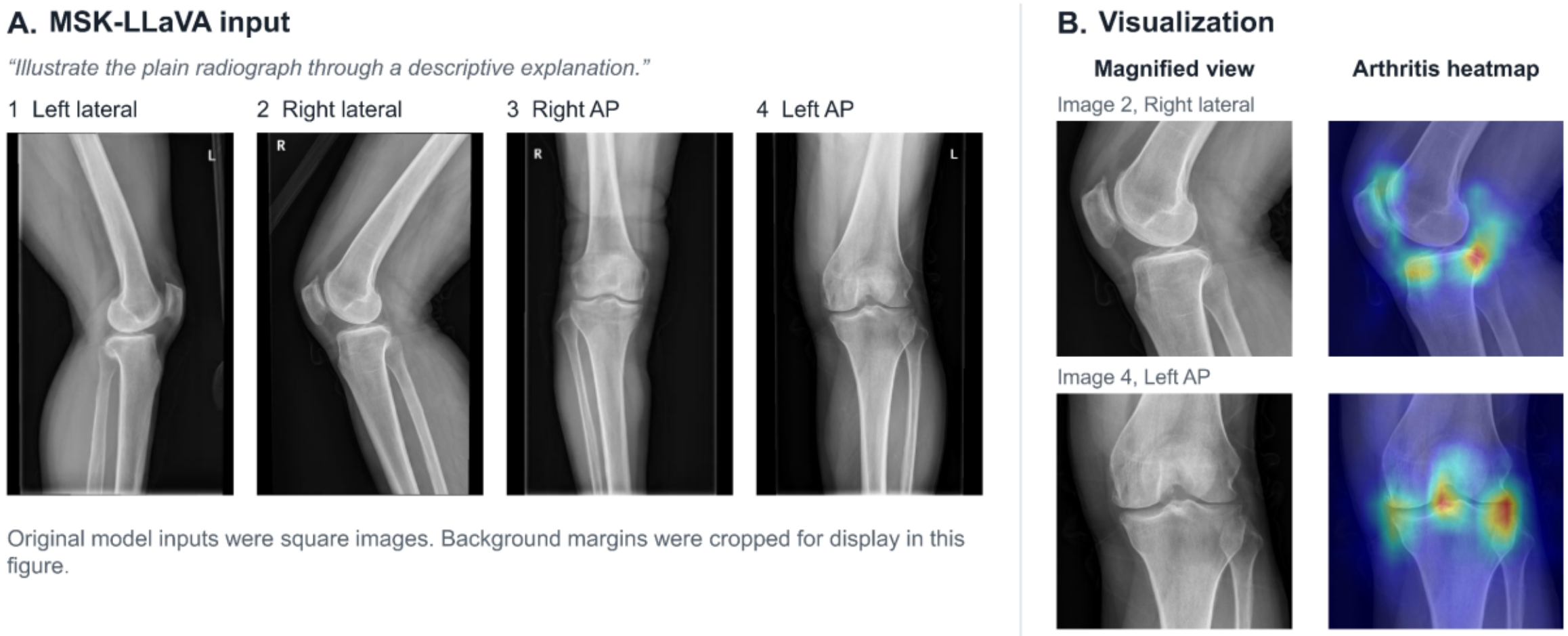

C. Report comparison

| | |
|---|---|
| **Reference standard** | **The knees demonstrate degenerative changes consistent with osteoarthritis.** No evidence of fracture, soft tissue swelling, joint effusion, osteoporosis or osteopenia, foreign body, prosthetic hardware, or joint dislocation is observed in either knee. |
| **MSK-LLaVA**<br>Bilateral OA reported | **Osteoarthritis is present in both the right and left knee joints.** No acute fracture is identified in either knee. |
| **MedGemma 1.5 (4B)**<br>OA not reported | Left knee: **The medial and lateral joint spaces are preserved.** No fracture or dislocation is identified. There is no significant joint effusion. The patella is normally located and oriented. No osseous lesions are seen. Soft tissues appear unremarkable. Right knee: **The medial and lateral joint spaces are preserved.** No fracture or dislocation is identified. There is no significant joint effusion. The patella is normally located and oriented. No osseous lesions are seen. Soft tissues appear unremarkable. |
| **MedGemma (27B)**<br>OA not reported | No acute fracture, dislocation, or significant osseous abnormality is identified. **Joint spaces appear preserved.** Soft tissues are unremarkable. |
| **Lingshu (32B)**<br>OA not reported | Findings: **The distal femur, proximal tibia, knee joint space, and patella appear normal bilaterally.**<br>There is no fracture-dislocation or bone destruction.<br>No significant soft tissue swelling is seen. |
| **Fleming-VL (38B)**<br>OA explicitly denied | The radiographs of the left and right knees show no acute fractures or dislocations. **The joint spaces are preserved, with no evidence of significant osteoarthritis such as subchondral sclerosis or large osteophytes. There is mild degenerative change noted in both knee joints,** characterized by slight narrowing of the medial compartment joint space on the lateral views. No soft tissue abnormalities are visible around either knee. |
| **MedDr (40B)**<br>OA not reported | The right knee shows a large, well-corticated ossific density in the suprapatellar region. **The left knee is unremarkable.** |
| **Gemma 4 (31B)**<br>OA not reported | Bilateral knee radiographs demonstrate normal alignment and **preserved joint spaces**. There is no evidence of acute fracture, dislocation, or suspicious osseous lesion. No joint effusion or significant soft-tissue abnormality is identified. |
| **Qwen3.8 (27B)**<br>OA not reported | Bilateral knee radiographs in anteroposterior and lateral projections demonstrate normal alignment of the femorotibial and patellofemoral joints without evidence of acute fracture or dislocation. **The joint spaces are preserved without significant narrowing, and there is no periarticular osteophyte formation, subchondral sclerosis, or cystic change to suggest advanced degenerative disease.** The osseous structures, including the distal femora, proximal tibiae, fibulae, and patellae, show intact cortices and normal trabecular pattern without focal lytic or sclerotic lesions. No joint effusion, soft-tissue swelling, or calcification is identified. No acute osseous or soft-tissue abnormality is seen. |

Radiographs of a 57-year-old female patient show bilateral knee osteoarthritis. (A) Anteroposterior and lateral radiographs of both knees used as model inputs. Background margins were cropped for display. (B) Magnified right lateral and left anteroposterior views with the corresponding arthritis heatmaps. (C) Reference and model-generated reports. MSK-LLaVA correctly reported bilateral knee osteoarthritis. Fleming-VL described mild bilateral degenerative changes with slight medial joint space narrowing, while also describing preserved joint spaces and no significant osteoarthritis. The other six comparator models did not identify osteoarthritis in either knee. Blue text highlights osteoarthritis or its radiographic features. Orange text highlights statements describing normal joint findings or negating osteoarthritis.

## Figure 5. Report generation for shoulder calcification.

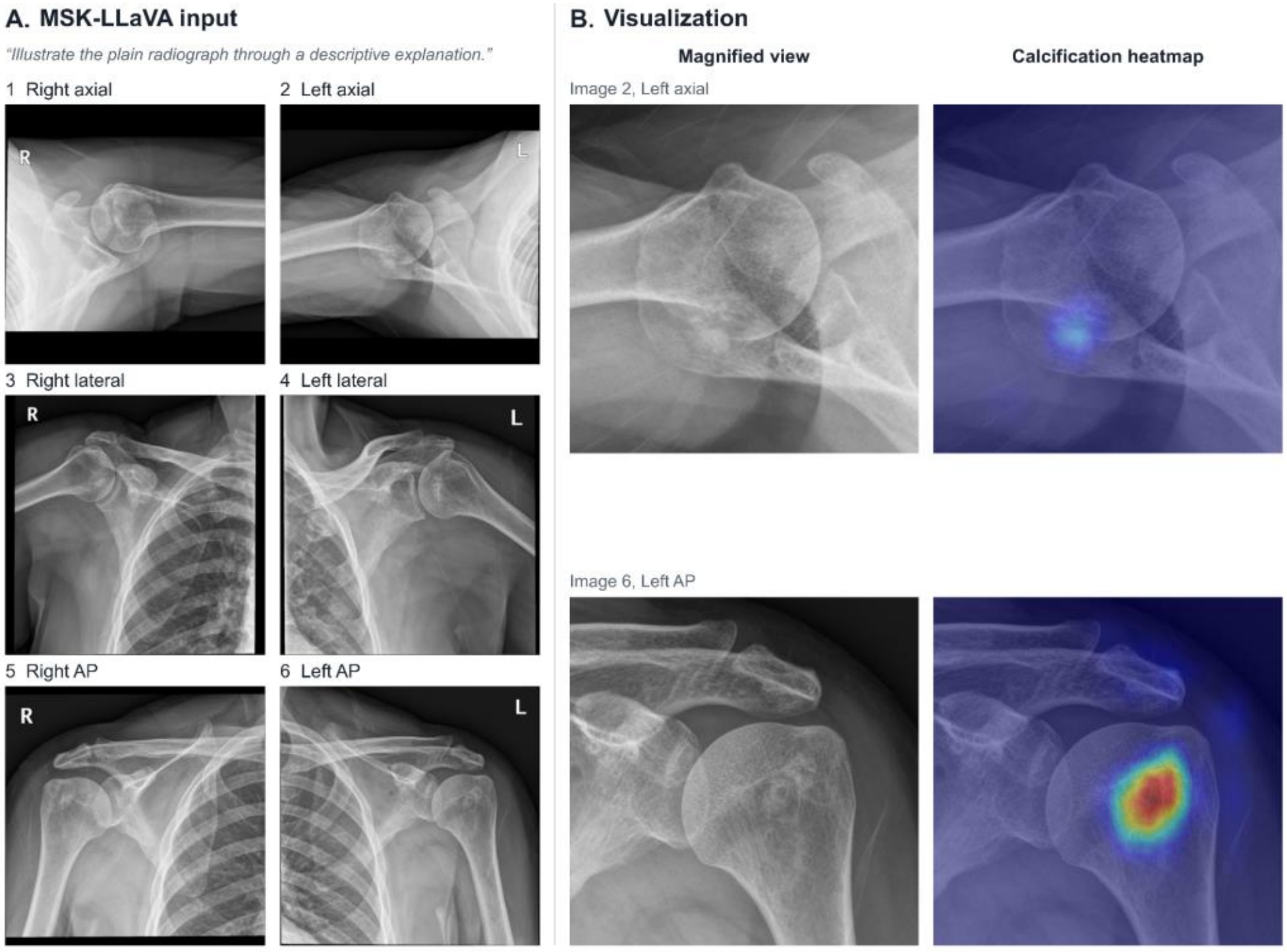

### C. Report comparison

| | |
|---|---|
| **Reference standard** | **Amorphous calcification is present about the left humeral greater tuberosity.** There is acromioclavicular joint arthrosis in both shoulders. No evidence of fracture, soft tissue swelling, joint effusion, osteoporosis or osteopenia, foreign body, prosthetic hardware, or joint dislocation in either shoulder. |
| **MSK-LLaVA**<br>Calcification reported | **A calcific deposit is identified in the left shoulder, consistent with calcific tendinitis.** No acute fracture is identified in either shoulder. |
| **MedGemma 1.5 (4B)**<br>Calcification not reported | FINDINGS: The right shoulder demonstrates normal alignment and joint spaces. There is no evidence of fracture or dislocation. The left shoulder demonstrates normal alignment and joint spaces. There is no evidence of fracture or dislocation. IMPRESSION: No acute abnormality. |
| **MedGemma (27B)**<br>Calcification not reported | No acute fracture, dislocation, or significant osseous abnormality is identified. The glenohumeral joint spaces appear preserved bilaterally. No significant soft tissue abnormality is seen. |
| **Lingshu (32B)**<br>Calcification not reported | Findings: The right shoulder demonstrates degenerative changes with narrowing of acromiohumeral interval, subchondral sclerosis at glenohumeral joint surfaces, and multiple loose bodies within the joint space.<br>The left shoulder is not significantly different from prior study with mild osteoarthritic change including spurring along humeral head articular margins and slight flattening of greater tuberosity suggestive of rotator cuff pathology. No definite loose body identified on this side however evaluation limited due to overlying soft tissues. There has been no significant change since recent comparison studies dated ___ except for possible increase in size of largest loose body seen anteriorly adjacent to coracoid process which measures approximately 9 mm compared to previous measurement of 7 mm. This could be positional but short-term followup imaging may be considered if clinically indicated. Impression: As above.. |
| **Fleming-VL (38B)**<br>Calcification not reported | The radiographs of the right and left shoulders demonstrate bilateral glenohumeral joints with no acute fractures or dislocations. The humeral heads are well-seated within the glenoid fossae, indicating normal joint alignment. There is evidence of degenerative changes in both shoulder girdles, including osteophyte formation at the acromioclavicular (AC) joints bilaterally. These findings suggest chronic AC joint arthropathy. **No significant soft tissue abnormalities such as calcifications or masses are noted on these images.**<br><br>In summary:<br>- Bilateral glenohumeral joints: Normal alignment without acute injury.<br>- Acromioclavicular joints: Degenerative changes consistent with chronic arthropathy.<br>- Soft tissues: Unremarkable for any acute pathology. |
| **MedDr (40B)**<br>Calcification not reported | The image presents a black and white X-ray of the right shoulder, captured in an AP view. The bones are clearly visible, with the humerus, scapula, and clavicle all discernible. There is no evidence of any fractures or other abnormalities in the image. |
| **Gemma 4 (31B)**<br>Calcification not reported | Right shoulder radiographs demonstrate dorsal luxation of the glenohumeral joint. No acute fracture or osseous lesion is identified in the right humerus or scapula. Left shoulder radiographs show normal alignment of the glenohumeral joint with preserved joint space and no evidence of acute fracture or dislocation. Soft tissues are unremarkable bilaterally. |
| **Qwen3.8 (27B)**<br>Calcification not reported | The right and left shoulder radiographs (AP, lateral, and axial views) demonstrate concentric articulation of both humeral heads within the glenoid fossae without evidence of dislocation or subluxation. The acromioclavicular and glenohumeral joint spaces are maintained, and the proximal humeri, scapulae, and clavicles show intact cortical margins without acute fracture, malalignment, or focal lytic or sclerotic osseous lesion. **No periosteal reaction, joint effusion, or abnormal soft-tissue calcification or mass is identified.** There is no acute osseous or articular abnormality. |

Radiographs of a 63-year-old female patient show an amorphous calcific deposit adjacent to the left greater tuberosity, consistent with calcific tendinitis. (A) Axial, lateral, and anteroposterior radiographs of both shoulders used as model inputs. (B) Magnified left axial and anteroposterior views with the corresponding calcification heatmaps. (C) Reference and model-generated reports. MSK-LLaVA described left shoulder calcification consistent with calcific tendinitis but omitted the precise location of the deposit. None of the seven comparator models reported the deposit. Blue text highlights descriptions of left shoulder calcification. Orange text highlights statements negating calcification.

**Figure 6. Report-generation failure in the external test set.**

**A. MSK-LLaVA input**

*"Illustrate the plain radiograph through a descriptive explanation."*

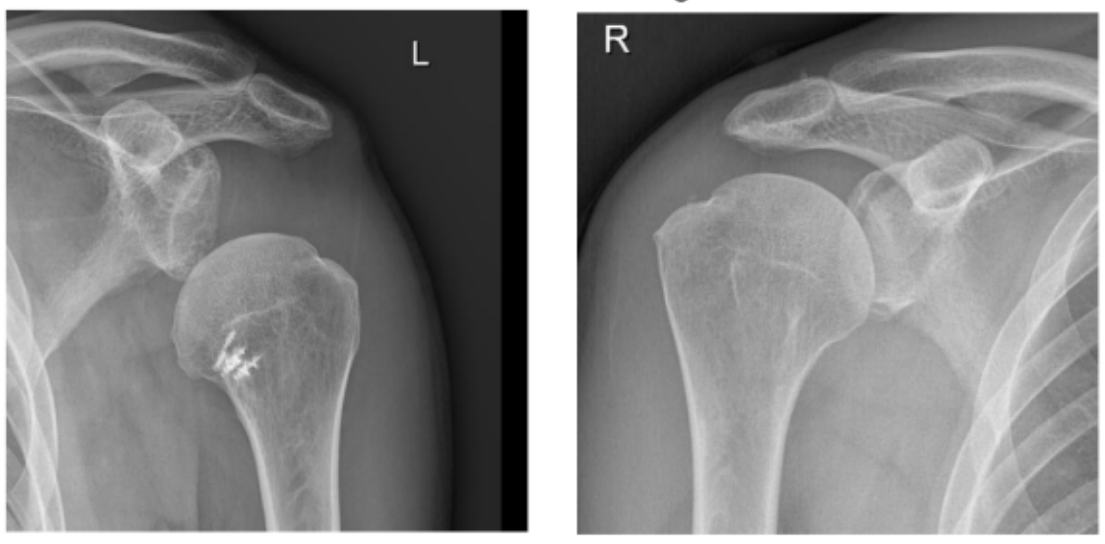

**B. Report comparison**

| Model | Report |
|---|---|
| **Reference standard** | **Anterior dislocation of left shoulder.** Mild glenohumeral osteoarthritis, both. |
| **MSK-LLaVA** Dislocation not reported | A Hill-Sachs lesion is identified in the left humerus. No Hill-Sachs lesion is identified in the right humerus. No acute fracture is identified in either shoulder. |
| **MedGemma 1.5 (4B)** Dislocation denied | Left shoulder: Status post surgical fixation of the greater tuberosity with multiple screws. **No evidence of acute fracture or dislocation.** Glenohumeral joint space is preserved. Acromioclavicular joint is unremarkable. Soft tissues are unremarkable. Right shoulder: No acute abnormality identified. The glenohumeral joint space is preserved. Acromioclavicular joint is unremarkable. Soft tissues are unremarkable. |
| **MedGemma (27B)** Dislocation denied | **No acute fracture, dislocation, or significant osseous abnormality is identified in either the left or right shoulder radiographs.** Joint spaces appear preserved. Soft tissues are unremarkable. |
| **Lingshu (32B)** Dislocation not reported | Findings: The patient is status post left proximal humeral fracture with fixation hardware present. There has been no significant change since prior study from ___ including callus formation at the surgical neck region, but lack of bony union. The right shoulder demonstrates degenerative changes within the acromioclavicular joint. No acute osseous abnormalities identified. No fractures seen. There is mild scoliosis convex to the right centered around T4-5 level. Both lungs appear clear bilaterally on this nondedicated exam. Visualized soft tissues are unremarkable. Impression: Status post left proximal humerus fracture repair with stable appearance compared to previous imaging studies; however, there remains nonunion between the head fragment and shaft fragments. . |
| **Fleming-VL (38B)** Dislocation denied | The radiographs show the left shoulder with a surgical screw fixation in place, indicating prior surgery. The bone structure around the humeral head and proximal shaft appears intact **without any visible fractures or dislocations.** There is no evidence of acute bony abnormalities on either side. Soft tissue shadows are normal bilaterally, and there are no signs of significant swelling or abnormal calcifications. |
| **MedDr (40B)** Dislocation not reported | The right humeral head is not well seen. There are no other significant findings |
| **Gemma 4 (31B)** Bilateral dislocation reported | **Bilateral anterior-inferior glenohumeral dislocations are present.** No acute fractures are identified. A radiopaque surgical anchor is noted in the superior aspect of the left humeral head. The acromioclavicular joints are aligned. No other osseous lesions or significant soft-tissue abnormalities are seen. |
| **Qwen3.8 (27B)** Dislocation denied | Bilateral anteroposterior shoulder radiographs demonstrate **no acute fracture or dislocation.** The glenohumeral and acromioclavicular joints are congruent with preserved joint spaces, and the humeral heads are appropriately seated within the glenoid fossae. No lytic or sclerotic osseous lesion is identified. The left proximal humeral head contains a small cluster of metallic surgical hardware consistent with prior anchor or screw fixation. No significant soft-tissue abnormality is seen. |

Radiographs of a 39-year-old male patient show a left anterior shoulder dislocation. (A) Anteroposterior radiographs of both shoulders used as model inputs. (B) Reference and model-generated reports. MSK-LLaVA reported a left humeral Hill-Sachs lesion, a finding associated with anterior shoulder dislocation, but did not identify the dislocation itself or mention the radiopaque surgical anchors in the left humeral head. Gemma 4 reported bilateral anterior-inferior glenohumeral dislocations, identifying the left-sided dislocation but incorrectly reporting a right-sided dislocation. MedGemma 1.5 (4B), MedGemma (27B), Fleming-VL, and Qwen3.8 explicitly denied dislocation. Qwen3.8 described the left proximal humeral surgical hardware. Lingshu described left proximal humeral fracture nonunion without mentioning dislocation. MedDr stated that the right humeral head was not well seen and reported no other significant findings. Blue and green text highlight statements describing dislocation, and orange text highlights statements explicitly negating dislocation.

# Supplemental Tables.

## Supplemental Table 1. Detailed Imaging Characteristics of the Internal Musculoskeletal Radiograph Dataset

| Characteristic | No. (%) |
|---|---|
| **No. of radiographs per examination** | |
| **1** | 554,058 (59.65) |
| **2** | 327,862 (35.30) |
| **3** | 6,070 (0.65) |
| **4** | 29,750 (3.20) |
| **5** | 10,687 (1.15) |
| **≥6** | 412 (0.04) |
| **Body part** | |
| **Knee** | 303,718 (21.63) |
| **Hip** | 157,705 (11.23) |
| **Hand** | 153,516 (10.93) |
| **Wrist** | 145,296 (10.35) |
| **Ankle** | 136,773 (9.74) |
| **Shoulder** | 126,880 (9.04) |
| **Foot** | 119,639 (8.52) |
| **Elbow** | 97,036 (6.91) |
| **Lower leg** | 39,337 (2.80) |
| **Pelvis** | 37,889 (2.70) |
| **Femur** | 30,299 (2.16) |
| **Forearm** | 21,725 (1.55) |
| **Upper arm** | 18,569 (1.32) |
| **Clavicle** | 9,909 (0.71) |
| **Scapula** | 5,814 (0.41) |
| **Laterality** | |
| **Right** | 613,041 (43.66) |
| **Left** | 591,171 (42.10) |
| **Not assigned** | 199,893 (14.24) |

*Note.*— The distribution of the number of radiographs per examination was calculated at the examination level. Body-part and laterality distributions were calculated at the radiograph level.

## Supplemental Table 2. Distribution of Excluded Radiographs by Anatomic Region and Protocol Body Part

| Anatomic Region | Protocol Body-Part Category | No. of Excluded Radiographs (%) |
|---|---|---|
| ***Spine*** | *Subtotal* | *984,016 (47.8)* |
| | Lumbosacral spine | 378,345 (18.4) |
| | Cervical spine | 291,537 (14.2) |
| | Lumbar spine | 157,314 (7.6) |
| | Whole spine | 63,835 (3.1) |
| | Thoracic spine | 39,608 (1.9) |
| | Dorsolumbar spine | 28,275 (1.4) |
| | Thoracolumbar spine | 25,102 (1.2) |
| ***Lower extremity*** | *Subtotal* | *759,156 (36.9)* |
| | Knee | 528,182 (25.7) |
| | Hip | 74,314 (3.6) |
| | Foot | 59,360 (2.9) |
| | Lower leg | 56,758 (2.8) |
| | Ankle | 36,876 (1.8) |
| | Femur | 3,666 (0.2) |
| ***Upper extremity*** | *Subtotal* | *228,948 (11.1)* |
| | Shoulder | 163,318 (7.9) |
| | Scapula | 39,716 (1.9) |
| | Wrist | 15,554 (0.8) |
| | Hand | 4,425 (0.2) |
| | Clavicle | 4,328 (0.2) |
| | Elbow | 1,258 (0.1) |
| | Upper arm (humerus) | 205 (<0.1) |
| | Forearm | 144 (<0.1) |
| ***Chest wall*** | *Subtotal* | *38,795 (1.9)* |
| | Rib | 35,769 (1.7) |
| | Sternum | 3,026 (0.1) |
| ***Pelvis and sacroiliac joint*** | *Subtotal* | *12,792 (0.6)* |
| | Pelvis | 10,914 (0.5) |
| | Sacroiliac joint | 1,878 (0.1) |
| ***Other*** | *Subtotal* | *34,092 (1.7)* |
| | Teleradiography | 30,670 (1.5) |
| | Miscellaneous | 3,421 (0.2) |
| | External image input (CR), body part unavailable | 1 (<0.1) |
| **Total** | | 2,057,799 (100.0) |

## Supplemental Table 3. Class Distribution of the Internal Test Subset for Evaluating MSK-SigLIP and Comparator Image Encoders

| Target Finding | Zero-shot N | Positive, No. (%) | Negative, No. (%) | Linear Probe N | Positive, No. (%) | Negative, No. (%) |
|---|---|---|---|---|---|---|
| **Arthritis** | 200 | 37 (18.5) | 163 (81.5) | 200 | 37 (18.5) | 163 (81.5) |
| **Fracture** | 200 | 37 (18.5) | 163 (81.5) | 200 | 37 (18.5) | 163 (81.5) |
| **Joint dislocation** | 200 | 10 (5.0) | 190 (95.0) | 200 | 10 (5.0) | 190 (95.0) |
| **Osteopenia** | 200 | 38 (19.0) | 162 (81.0) | 200 | 38 (19.0) | 162 (81.0) |
| **Prosthetic hardware** | 200 | 15 (7.5) | 185 (92.5) | 200 | 15 (7.5) | 185 (92.5) |
| **Soft-tissue swelling** | 200 | 15 (7.5) | 185 (92.5) | 200 | 15 (7.5) | 185 (92.5) |

The internal test subset was used for zero-shot and linear-probe evaluations of the encoders.

## Supplemental Table 4. Class Distribution of the External Test Sets for Evaluating MSK-SigLIP and Comparator Image Encoders

| Dataset | Target Finding | Zero-shot N | Positive, No. (%) | Negative, No. (%) | Linear Probe N | Positive, No. (%) | Negative, No. (%) |
|---|---|---|---|---|---|---|---|
| **MURA (1)** | Abnormality | 1,199 | 538 (44.9) | 661 (55.1) | 1,199 | 538 (44.9) | 661 (55.1) |
| **FracAtlas (2)** | Fracture | 4,083 | 717 (17.6) | 3,366 (82.4) | 409 | 72 (17.6) | 337 (82.4) |
| **LERA (3)** | Abnormality | 168 | 95 (56.5) | 73 (43.5) | 17 | 10 (58.8) | 7 (41.2) |
| **GRAZPEDWRI-DX (4)** | Cast | 20,327 | 5,776 (28.4) | 14,551 (71.6) | 2,006 | 546 (27.2) | 1,460 (72.8) |
| | Fracture | 20,327 | 13,550 (66.7) | 6,777 (33.3) | 2,006 | 1,326 (66.1) | 680 (33.9) |
| | Metal | 20,327 | 708 (3.5) | 19,619 (96.5) | 2,006 | 56 (2.8) | 1,950 (97.2) |
| | Osteopenia | 20,327 | 2,473 (12.2) | 17,854 (87.8) | 2,006 | 235 (11.7) | 1,771 (88.3) |
| **BTXRD (5)** | Any tumor | 3,746 | 1,867 (49.8) | 1,879 (50.2) | 375 | 187 (49.9) | 188 (50.1) |
| **BTXRD tumor-positive subset** | Malignancy | 1,867 | 342 (18.3) | 1,525 (81.7) | 187 | 34 (18.2) | 153 (81.8) |

*MURA and LERA were evaluated after examination-level aggregation. Zero-shot evaluation used the full available external evaluation set, whereas linear-probe evaluation required fitting a classifier and was therefore performed on a held-out test subset; consequently, sample sizes differ between the two evaluation settings for some datasets.*

# Supplemental Table 5. Zero-Shot Classification Performance of MSK-SigLIP and Comparator Image Encoders on Internal and External Musculoskeletal Radiograph Datasets

| Dataset | Task | MSK-SigLIP | MedSigLIP | OpenAI CLIP | SigLIP2 |
|---|---|---|---|---|---|
| **Internal test subset** | Fracture | **0.87 (0.77, 0.94)** | 0.67 (0.56, 0.77) | 0.51 (0.40, 0.62) | 0.55 (0.44, 0.66) |
| | Arthritis | **0.81 (0.73, 0.88)** | 0.61 (0.50, 0.71) | 0.36 (0.27, 0.46) | 0.36 (0.26, 0.46) |
| | Joint dislocation | **0.97 (0.91, 1.00)** | 0.60 (0.45, 0.75) | 0.45 (0.29, 0.61) | 0.40 (0.22, 0.58) |
| | Osteopenia | **0.80 (0.71, 0.87)** | 0.70 (0.59, 0.79) | 0.50 (0.40, 0.59) | 0.44 (0.34, 0.55) |
| | Prosthetic hardware | 0.81 (0.66, 0.93) | **0.90 (0.81, 0.96)** | 0.84 (0.73, 0.93) | 0.66 (0.52, 0.79) |
| | Soft-tissue swelling | **0.83 (0.74, 0.91)** | 0.58 (0.44, 0.72) | 0.52 (0.36, 0.68) | 0.65 (0.52, 0.79) |
| **MURA** | Abnormality† | **0.83 (0.81, 0.86)** | 0.77 (0.74, 0.80) | 0.53 (0.50, 0.57) | 0.65 (0.62, 0.69) |
| **FracAtlas** | Fracture | **0.84 (0.82, 0.86)** | 0.61 (0.59, 0.63) | 0.55 (0.53, 0.58) | 0.54 (0.51, 0.56) |
| **LERA** | Abnormality† | **0.82 (0.76, 0.88)** | 0.78 (0.71, 0.84) | 0.53 (0.44, 0.62) | 0.51 (0.42, 0.60) |
| **GRAZPEDWRI-DX** | Fracture | **0.92 (0.92, 0.92)** | 0.68 (0.67, 0.69) | 0.52 (0.51, 0.53) | 0.64 (0.64, 0.65) |
| | Cast | 0.87 (0.86, 0.87) | **0.87 (0.87, 0.88)** | 0.59 (0.58, 0.60) | 0.70 (0.69, 0.71) |
| | Metal | 0.86 (0.84, 0.87) | 0.96 (0.96, 0.97) | **0.98 (0.97, 0.98)** | 0.91 (0.90, 0.92) |
| | Osteopenia | **0.68 (0.67, 0.69)** | 0.51 (0.50, 0.52) | 0.54 (0.53, 0.55) | 0.61 (0.60, 0.62) |
| **BTXRD** | Tumor | **0.92 (0.91, 0.93)** | 0.01 (0.01, 0.01) | 0.56 (0.54, 0.58) | 0.30 (0.28, 0.32) |
| **BTXRD tumor-positive subset** | Malignancy | 0.49 (0.46, 0.53) | **0.79 (0.76, 0.82)** | 0.50 (0.47, 0.53) | 0.71 (0.67, 0.74) |

*Note.— Data are AUC values, with 95% CIs in parentheses. Boldface indicates the highest AUC for each task based on unrounded values. †For MURA and LERA, radiograph-level prediction scores were averaged within each examination before performance metrics were calculated.*

# Supplemental Table 6. Linear-Probe Classification Performance of MSK-SigLIP and Comparator Image Encoders on Internal and External Musculoskeletal Radiograph Datasets

| Dataset | Task | MSK-SigLIP | MedSigLIP | RadioDINO | SKELEX | OpenAI CLIP | SigLIP2 | DINOv3 | DINOv3-CNN |
|---|---|---|---|---|---|---|---|---|---|
| | | | **Medical-domain models** | | | **General-domain models** | | | |
| **Internal test set** | Fracture | **0.90 (0.82, 0.96)** | 0.78 (0.70, 0.86) | 0.77 (0.69, 0.84) | 0.78 (0.70, 0.87) | 0.81 (0.73, 0.89) | 0.80 (0.71, 0.88) | 0.75 (0.66, 0.83) | 0.78 (0.70, 0.85) |
| | Arthritis | **0.84 (0.77, 0.91)** | 0.74 (0.64, 0.82) | 0.67 (0.58, 0.76) | 0.73 (0.63, 0.82) | 0.77 (0.68, 0.85) | 0.71 (0.61, 0.80) | 0.73 (0.64, 0.82) | 0.71 (0.61, 0.80) |
| | Joint dislocation | 0.67 (0.47, 0.84) | 0.71 (0.52, 0.87) | 0.46 (0.25, 0.68) | 0.66 (0.46, 0.84) | 0.70 (0.51, 0.87) | **0.72 (0.60, 0.83)** | 0.65 (0.45, 0.80) | 0.55 (0.39, 0.71) |
| | Osteopenia | **0.86 (0.80, 0.91)** | 0.76 (0.68, 0.84) | 0.70 (0.61, 0.79) | 0.83 (0.75, 0.89) | 0.75 (0.67, 0.83) | 0.76 (0.69, 0.84) | 0.77 (0.70, 0.85) | 0.73 (0.64, 0.81) |
| | Prosthetic hardware | 0.87 (0.73, 0.98) | 0.97 (0.91, 1.00) | 0.84 (0.71, 0.95) | 0.93 (0.84, 0.99) | 0.96 (0.87, 1.00) | 0.99 (0.96, 1.00) | **0.99 (0.98, 1.00)** | 0.97 (0.95, 0.99) |
| | Soft-tissue swelling | **0.94 (0.89, 0.97)** | 0.91 (0.84, 0.96) | 0.80 (0.70, 0.89) | 0.92 (0.87, 0.96) | 0.83 (0.71, 0.92) | 0.91 (0.85, 0.96) | 0.90 (0.83, 0.96) | 0.77 (0.64, 0.87) |
| **MURA** | Abnormality† | **0.91 (0.90, 0.93)** | 0.88 (0.85, 0.89) | 0.82 (0.80, 0.85) | 0.87 (0.85, 0.89) | 0.87 (0.85, 0.89) | 0.90 (0.88, 0.91) | 0.89 (0.87, 0.91) | 0.86 (0.84, 0.89) |
| **FracAtlas** | Fracture | **0.98 (0.96, 0.99)** | 0.93 (0.90, 0.96) | 0.87 (0.82, 0.92) | 0.92 (0.89, 0.95) | 0.89 (0.85, 0.94) | 0.93 (0.90, 0.96) | 0.94 (0.91, 0.97) | 0.91 (0.87, 0.95) |
| **LERA** | Abnormality† | 0.80 (0.54, 1.00) | **0.86 (0.64, 1.00)** | 0.41 (0.14, 0.70) | 0.74 (0.46, 0.97) | 0.66 (0.40, 0.91) | 0.73 (0.46, 0.94) | 0.80 (0.57, 0.97) | 0.66 (0.37, 0.90) |
| **GRAZPEDWRI-DX** | Fracture | **0.97 (0.96, 0.97)** | 0.91 (0.89, 0.92) | 0.86 (0.84, 0.87) | 0.90 (0.88, 0.91) | 0.89 (0.87, 0.90) | 0.91 (0.90, 0.92) | 0.93 (0.92, 0.94) | 0.89 (0.87, 0.90) |
| | Cast | 1.00 (1.00, 1.00) | 1.00 (1.00, 1.00) | 1.00 (1.00, 1.00) | 1.00 (1.00, 1.00) | 1.00 (1.00, 1.00) | 1.00 (1.00, 1.00) | **1.00 (1.00, 1.00)** | 1.00 (1.00, 1.00) |
| | Metal | 1.00 (0.99, 1.00) | 1.00 (1.00, 1.00) | 0.98 (0.95, 1.00) | 0.99 (0.98, 1.00) | 1.00 (1.00, 1.00) | **1.00 (1.00, 1.00)** | 1.00 (1.00, 1.00) | 0.99 (0.98, 1.00) |
| | Osteopenia | **0.97 (0.95, 0.98)** | 0.97 (0.95, 0.98) | 0.91 (0.89, 0.94) | 0.97 (0.95, 0.98) | 0.94 (0.92, 0.96) | 0.94 (0.93, 0.96) | 0.97 (0.95, 0.98) | 0.95 (0.93, 0.96) |
| **BTXRD** | Tumor | **0.95 (0.93, 0.97)** | 0.93 (0.90, 0.95) | 0.84 (0.79, 0.88) | 0.88 (0.84, 0.91) | 0.89 (0.85, 0.92) | 0.92 (0.89, 0.94) | 0.93 (0.90, 0.95) | 0.87 (0.84, 0.91) |
| **BTXRD tumor-positive subset** | Malignancy | 0.95 (0.90, 0.99) | 0.95 (0.91, 0.98) | 0.86 (0.78, 0.93) | 0.92 (0.85, 0.97) | 0.92 (0.86, 0.96) | **0.97 (0.94, 0.99)** | 0.94 (0.87, 0.98) | 0.85 (0.76, 0.93) |

*Note.— Data are AUC values, with 95% CIs in parentheses. Boldface indicates the highest AUC for each task based on unrounded values. †For MURA and LERA, radiograph-level prediction scores were averaged within each examination before performance metrics were calculated.*

Supplemental Table 7. Sensitivity, Specificity, and Confusion-Matrix Counts for Detection of Radiographic Findings by MSK-LLaVA and Comparator Vision–Language Models in the Internal Test Set

## Panel A. Sensitivity

| Model | Fracture | Soft-tissue swelling | Arthritis | Prosthetic hardware | Osteopenia | Joint dislocation |
|---|---|---|---|---|---|---|
| **MSK-LLaVA** | **0.86 (0.74, 0.97)** | **0.24 (0.06, 0.47)** | **0.54 (0.23, 0.77)** | **1.00 (1.00, 1.00)** | **0.33 (0.00, 1.00)** | **0.50 (0.00, 1.00)** |
| **MedGemma 1.5 (4B)** | 0.23 (0.09, 0.37) | 0.00 (0.00, 0.00) | 0.00 (0.00, 0.00) | **1.00 (1.00, 1.00)** | 0.00 (0.00, 0.00) | 0.00 (0.00, 0.00) |
| **MedGemma (27B)** | 0.20 (0.09, 0.34) | 0.00 (0.00, 0.00) | 0.08 (0.00, 0.23) | 0.88 (0.62, 1.00) | 0.00 (0.00, 0.00) | 0.25 (0.00, 0.75) |
| **Lingshu (32B)** | 0.40 (0.23, 0.57) | 0.06 (0.00, 0.18) | 0.46 (0.15, 0.69) | **1.00 (1.00, 1.00)** | 0.00 (0.00, 0.00) | 0.00 (0.00, 0.00) |
| **Fleming-VL (38B)** | 0.17 (0.06, 0.31) | 0.12 (0.00, 0.29) | 0.31 (0.08, 0.54) | **1.00 (1.00, 1.00)** | 0.00 (0.00, 0.00) | 0.00 (0.00, 0.00) |
| **MedDr (40B)** | 0.20 (0.09, 0.34) | 0.00 (0.00, 0.00) | 0.00 (0.00, 0.00) | 0.88 (0.62, 1.00) | 0.00 (0.00, 0.00) | 0.00 (0.00, 0.00) |
| **Gemma 4 (31B)** | 0.34 (0.17, 0.51) | 0.06 (0.00, 0.18) | 0.08 (0.00, 0.23) | **1.00 (1.00, 1.00)** | 0.00 (0.00, 0.00) | **0.50 (0.00, 1.00)** |
| **Qwen3.8 (27B)** | 0.14 (0.03, 0.26) | 0.00 (0.00, 0.00) | 0.00 (0.00, 0.00) | **1.00 (1.00, 1.00)** | 0.00 (0.00, 0.00) | **0.50 (0.00, 1.00)** |

## Panel B. Specificity

| Model | Fracture | Soft-tissue swelling | Arthritis | Prosthetic hardware | Osteopenia | Joint dislocation |
|---|---|---|---|---|---|---|
| **MSK-LLaVA** | 0.92 (0.85, 0.98) | 0.96 (0.92, 1.00) | 0.95 (0.91, 0.99) | 0.98 (0.95, 1.00) | 0.94 (0.89, 0.98) | **1.00 (1.00, 1.00)** |
| **MedGemma 1.5 (4B)** | 0.94 (0.88, 0.98) | **1.00 (1.00, 1.00)** | 0.97 (0.92, 1.00) | 0.96 (0.91, 0.99) | **1.00 (1.00, 1.00)** | **1.00 (1.00, 1.00)** |
| **MedGemma (27B)** | 0.92 (0.86, 0.98) | 0.94 (0.89, 0.99) | 0.98 (0.94, 1.00) | **1.00 (1.00, 1.00)** | **1.00 (1.00, 1.00)** | **1.00 (1.00, 1.00)** |
| **Lingshu (32B)** | 0.78 (0.68, 0.88) | 0.89 (0.82, 0.95) | 0.60 (0.49, 0.70) | 0.95 (0.89, 0.99) | 0.91 (0.85, 0.96) | 0.91 (0.84, 0.96) |
| **Fleming-VL (38B)** | 0.95 (0.89, 1.00) | 0.92 (0.86, 0.96) | 0.83 (0.75, 0.91) | 0.98 (0.95, 1.00) | 0.99 (0.97, 1.00) | 0.98 (0.95, 1.00) |
| **MedDr (40B)** | 0.86 (0.78, 0.94) | 0.94 (0.88, 0.99) | 0.90 (0.83, 0.95) | 0.93 (0.88, 0.98) | 0.94 (0.89, 0.98) | 0.96 (0.92, 0.99) |
| **Gemma 4 (31B)** | 0.95 (0.91, 1.00) | 0.86 (0.77, 0.93) | **1.00 (1.00, 1.00)** | 0.96 (0.91, 0.99) | **1.00 (1.00, 1.00)** | 0.82 (0.75, 0.90) |
| **Qwen3.8 (27B)** | **0.98 (0.95, 1.00)** | 0.96 (0.92, 1.00) | 0.98 (0.94, 1.00) | 0.95 (0.89, 0.99) | **1.00 (1.00, 1.00)** | 0.98 (0.95, 1.00) |

## Panel C. Confusion-Matrix Counts for Finding-Level Classification in the Internal Test Set

| Model | Finding | Positive | Negative | TP | TN | FP | FN |
|---|---|---|---|---|---|---|---|
| **MSK-LLaVA** | Fracture | 35 | 65 | 30 | 60 | 5 | 5 |
| | Soft-tissue swelling | 17 | 83 | 4 | 80 | 3 | 13 |
| | Arthritis | 13 | 87 | 7 | 83 | 4 | 6 |
| | Prosthetic hardware | 8 | 92 | 8 | 90 | 2 | 0 |
| | Osteopenia | 3 | 97 | 1 | 91 | 6 | 2 |
| | Joint dislocation | 4 | 96 | 2 | 96 | 0 | 2 |
| **MedGemma 1.5 (4B)** | Fracture | 35 | 65 | 8 | 61 | 4 | 27 |
| | Soft-tissue swelling | 17 | 83 | 0 | 83 | 0 | 17 |
| | Arthritis | 13 | 87 | 0 | 84 | 3 | 13 |
| | Prosthetic hardware | 8 | 92 | 8 | 88 | 4 | 0 |
| | Osteopenia | 3 | 97 | 0 | 97 | 0 | 3 |
| | Joint dislocation | 4 | 96 | 0 | 96 | 0 | 4 |
| **MedGemma (27B)** | Fracture | 35 | 65 | 7 | 60 | 5 | 28 |
| | Soft-tissue swelling | 17 | 83 | 0 | 78 | 5 | 17 |
| | Arthritis | 13 | 87 | 1 | 85 | 2 | 12 |
| | Prosthetic hardware | 8 | 92 | 7 | 92 | 0 | 1 |
| | Osteopenia | 3 | 97 | 0 | 97 | 0 | 3 |
| | Joint dislocation | 4 | 96 | 1 | 96 | 0 | 3 |
| **Lingshu (32B)** | Fracture | 35 | 65 | 14 | 51 | 14 | 21 |
| | Soft-tissue swelling | 17 | 83 | 1 | 74 | 9 | 16 |
| | Arthritis | 13 | 87 | 6 | 52 | 35 | 7 |
| | Prosthetic hardware | 8 | 92 | 8 | 87 | 5 | 0 |
| | Osteopenia | 3 | 97 | 0 | 88 | 9 | 3 |
| | Joint dislocation | 4 | 96 | 0 | 87 | 9 | 4 |
| **Fleming-VL (38B)** | Fracture | 35 | 65 | 6 | 62 | 3 | 29 |
| | Soft-tissue swelling | 17 | 83 | 2 | 76 | 7 | 15 |
| | Arthritis | 13 | 87 | 4 | 72 | 15 | 9 |
| | Prosthetic hardware | 8 | 92 | 8 | 90 | 2 | 0 |
| | Osteopenia | 3 | 97 | 0 | 96 | 1 | 3 |
| | Joint dislocation | 4 | 96 | 0 | 94 | 2 | 4 |
| **MedDr (40B)** | Fracture | 35 | 65 | 7 | 56 | 9 | 28 |
| | Soft-tissue swelling | 17 | 83 | 0 | 78 | 5 | 17 |
| | Arthritis | 13 | 87 | 0 | 78 | 9 | 13 |
| | Prosthetic hardware | 8 | 92 | 7 | 86 | 6 | 1 |
| | Osteopenia | 3 | 97 | 0 | 91 | 6 | 3 |
| | Joint dislocation | 4 | 96 | 0 | 92 | 4 | 4 |
| **Gemma 4 (31B)** | Fracture | 35 | 65 | 12 | 62 | 3 | 23 |
| | Soft-tissue swelling | 17 | 83 | 1 | 71 | 12 | 16 |
| | Arthritis | 13 | 87 | 1 | 87 | 0 | 12 |
| | Prosthetic hardware | 8 | 92 | 8 | 88 | 4 | 0 |
| | Osteopenia | 3 | 97 | 0 | 97 | 0 | 3 |
| | Joint dislocation | 4 | 96 | 2 | 79 | 17 | 2 |
| **Qwen3.8 (27B)** | Fracture | 35 | 65 | 5 | 64 | 1 | 30 |
| | Soft-tissue swelling | 17 | 83 | 0 | 80 | 3 | 17 |
| | Arthritis | 13 | 87 | 0 | 85 | 2 | 13 |
| | Prosthetic hardware | 8 | 92 | 8 | 87 | 5 | 0 |
| | Osteopenia | 3 | 97 | 0 | 97 | 0 | 3 |
| | Joint dislocation | 4 | 96 | 2 | 94 | 2 | 2 |

*Note.—The internal test set included 100 examinations. Data are sensitivity or specificity, with 95% CIs in parentheses. In panels A and B, boldface indicates the highest value for each finding, including ties. Findings were extracted from model-generated reports and compared with the expert reference standard. TP = true positive; TN = true negative; FP = false positive; FN = false negative.*

# Supplemental Table 8. Sensitivity, Specificity, and Confusion-Matrix Counts for Detection of Radiographic Findings by MSK-LLaVA and Comparator Vision–Language Models in the External Test Set

## Panel A. Sensitivity

| Model | Fracture | Soft-tissue swelling | Arthritis | Prosthetic hardware | Osteopenia | Joint dislocation |
|---|---|---|---|---|---|---|
| **MSK-LLaVA** | **0.96 (0.89, 1.00)** | **0.43 (0.14, 0.71)** | **0.71 (0.50, 0.93)** | 0.88 (0.62, 1.00) | **1.00 (1.00, 1.00)** | **0.33 (0.00, 1.00)** |
| **MedGemma 1.5 (4B)** | 0.13 (0.04, 0.23) | 0.00 (0.00, 0.00) | 0.00 (0.00, 0.00) | **1.00 (1.00, 1.00)** | 0.00 (0.00, 0.00) | 0.00 (0.00, 0.00) |
| **MedGemma (27B)** | 0.19 (0.09, 0.30) | 0.00 (0.00, 0.00) | 0.00 (0.00, 0.00) | 0.25 (0.00, 0.62) | 0.00 (0.00, 0.00) | 0.00 (0.00, 0.00) |
| **Lingshu (32B)** | 0.32 (0.19, 0.45) | 0.14 (0.00, 0.43) | 0.36 (0.14, 0.64) | **1.00 (1.00, 1.00)** | 0.00 (0.00, 0.00) | 0.00 (0.00, 0.00) |
| **Fleming-VL (38B)** | 0.04 (0.00, 0.11) | 0.00 (0.00, 0.00) | 0.43 (0.14, 0.71) | **1.00 (1.00, 1.00)** | 0.00 (0.00, 0.00) | 0.00 (0.00, 0.00) |
| **MedDr (40B)** | 0.26 (0.13, 0.38) | 0.00 (0.00, 0.00) | 0.21 (0.00, 0.43) | 0.75 (0.38, 1.00) | 0.00 (0.00, 0.00) | 0.00 (0.00, 0.00) |
| **Gemma 4 (31B)** | 0.17 (0.06, 0.30) | 0.00 (0.00, 0.00) | 0.00 (0.00, 0.00) | **1.00 (1.00, 1.00)** | 0.00 (0.00, 0.00) | **0.33 (0.00, 1.00)** |
| **Qwen3.8 (27B)** | 0.11 (0.02, 0.19) | 0.00 (0.00, 0.00) | 0.07 (0.00, 0.21) | **1.00 (1.00, 1.00)** | 0.00 (0.00, 0.00) | 0.00 (0.00, 0.00) |

## Panel B. Specificity

| Model | Fracture | Soft-tissue swelling | Arthritis | Prosthetic hardware | Osteopenia | Joint dislocation |
|---|---|---|---|---|---|---|
| **MSK-LLaVA** | 0.93 (0.86, 0.99) | 0.98 (0.95, 1.00) | 0.96 (0.92, 0.99) | **1.00 (1.00, 1.00)** | 0.91 (0.86, 0.96) | **1.00 (1.00, 1.00)** |
| **MedGemma 1.5 (4B)** | 0.96 (0.90, 1.00) | **0.99 (0.97, 1.00)** | 0.98 (0.95, 1.00) | 0.95 (0.91, 0.99) | **1.00 (1.00, 1.00)** | **1.00 (1.00, 1.00)** |
| **MedGemma (27B)** | 0.81 (0.71, 0.90) | 0.98 (0.95, 1.00) | **1.00 (1.00, 1.00)** | **1.00 (1.00, 1.00)** | **1.00 (1.00, 1.00)** | 0.93 (0.88, 0.97) |
| **Lingshu (32B)** | 0.88 (0.80, 0.96) | 0.89 (0.83, 0.94) | 0.63 (0.53, 0.72) | 0.91 (0.85, 0.95) | 0.83 (0.77, 0.90) | 0.87 (0.80, 0.92) |
| **Fleming-VL (38B)** | **1.00 (1.00, 1.00)** | 0.94 (0.89, 0.98) | 0.84 (0.77, 0.91) | 0.95 (0.91, 0.99) | **1.00 (1.00, 1.00)** | 0.97 (0.94, 1.00) |
| **MedDr (40B)** | 0.93 (0.86, 0.99) | 0.95 (0.91, 0.99) | 0.96 (0.92, 0.99) | 0.92 (0.86, 0.96) | 0.99 (0.97, 1.00) | 0.88 (0.81, 0.93) |
| **Gemma 4 (31B)** | 0.99 (0.96, 1.00) | 0.85 (0.79, 0.92) | **1.00 (1.00, 1.00)** | 0.94 (0.88, 0.98) | **1.00 (1.00, 1.00)** | 0.78 (0.70, 0.85) |
| **Qwen3.8 (27B)** | **1.00 (1.00, 1.00)** | 0.94 (0.90, 0.98) | 0.99 (0.97, 1.00) | 0.94 (0.90, 0.98) | 0.99 (0.97, 1.00) | 0.96 (0.91, 0.99) |

## Panel C. Confusion-Matrix Counts for Finding-Level Classification in the External Test Set

| Model | Finding | Positive | Negative | TP | TN | FP | FN |
|---|---|---|---|---|---|---|---|
| **MSK-LLaVA** | Fracture | 47 | 69 | 45 | 64 | 5 | 2 |
| | Soft-tissue swelling | 7 | 109 | 3 | 107 | 2 | 4 |
| | Arthritis | 14 | 102 | 10 | 98 | 4 | 4 |
| | Prosthetic hardware | 8 | 108 | 7 | 108 | 0 | 1 |
| | Osteopenia | 3 | 113 | 3 | 103 | 10 | 0 |
| | Joint dislocation | 3 | 113 | 1 | 113 | 0 | 2 |
| **MedGemma 1.5 (4B)** | Fracture | 47 | 69 | 6 | 66 | 3 | 41 |
| | Soft-tissue swelling | 7 | 109 | 0 | 108 | 1 | 7 |
| | Arthritis | 14 | 102 | 0 | 100 | 2 | 14 |
| | Prosthetic hardware | 8 | 108 | 8 | 103 | 5 | 0 |
| | Osteopenia | 3 | 113 | 0 | 113 | 0 | 3 |
| | Joint dislocation | 3 | 113 | 0 | 113 | 0 | 3 |
| **MedGemma (27B)** | Fracture | 47 | 69 | 9 | 56 | 13 | 38 |
| | Soft-tissue swelling | 7 | 109 | 0 | 107 | 2 | 7 |
| | Arthritis | 14 | 102 | 0 | 102 | 0 | 14 |
| | Prosthetic hardware | 8 | 108 | 2 | 108 | 0 | 6 |
| | Osteopenia | 3 | 113 | 0 | 113 | 0 | 3 |
| | Joint dislocation | 3 | 113 | 0 | 105 | 8 | 3 |
| **Lingshu (32B)** | Fracture | 47 | 69 | 15 | 61 | 8 | 32 |
| | Soft-tissue swelling | 7 | 109 | 1 | 97 | 12 | 6 |
| | Arthritis | 14 | 102 | 5 | 64 | 38 | 9 |
| | Prosthetic hardware | 8 | 108 | 8 | 98 | 10 | 0 |
| | Osteopenia | 3 | 113 | 0 | 94 | 19 | 3 |
| | Joint dislocation | 3 | 113 | 0 | 98 | 15 | 3 |
| **Fleming-VL (38B)** | Fracture | 47 | 69 | 2 | 69 | 0 | 45 |
| | Soft-tissue swelling | 7 | 109 | 0 | 102 | 7 | 7 |
| | Arthritis | 14 | 102 | 6 | 86 | 16 | 8 |
| | Prosthetic hardware | 8 | 108 | 8 | 103 | 5 | 0 |
| | Osteopenia | 3 | 113 | 0 | 113 | 0 | 3 |
| | Joint dislocation | 3 | 113 | 0 | 110 | 3 | 3 |
| **MedDr (40B)** | Fracture | 47 | 69 | 12 | 64 | 5 | 35 |
| | Soft-tissue swelling | 7 | 109 | 0 | 104 | 5 | 7 |
| | Arthritis | 14 | 102 | 3 | 98 | 4 | 11 |
| | Prosthetic hardware | 8 | 108 | 6 | 99 | 9 | 2 |
| | Osteopenia | 3 | 113 | 0 | 112 | 1 | 3 |
| | Joint dislocation | 3 | 113 | 0 | 99 | 14 | 3 |
| **Gemma 4 (31B)** | Fracture | 47 | 69 | 8 | 68 | 1 | 39 |
| | Soft-tissue swelling | 7 | 109 | 0 | 93 | 16 | 7 |
| | Arthritis | 14 | 102 | 0 | 102 | 0 | 14 |
| | Prosthetic hardware | 8 | 108 | 8 | 101 | 7 | 0 |
| | Osteopenia | 3 | 113 | 0 | 113 | 0 | 3 |
| | Joint dislocation | 3 | 113 | 1 | 88 | 25 | 2 |
| **Qwen3.8 (27B)** | Fracture | 47 | 69 | 5 | 69 | 0 | 42 |
| | Soft-tissue swelling | 7 | 109 | 0 | 103 | 6 | 7 |
| | Arthritis | 14 | 102 | 1 | 101 | 1 | 13 |
| | Prosthetic hardware | 8 | 108 | 8 | 102 | 6 | 0 |
| | Osteopenia | 3 | 113 | 0 | 112 | 1 | 3 |
| | Joint dislocation | 3 | 113 | 0 | 108 | 5 | 3 |

*Note.—All models were evaluated on the same external test set of 116 examinations. Data are sensitivity or specificity, with 95% CIs in parentheses. In panels A and B, boldface indicates the highest value for each finding, including ties. Findings extracted from model-generated reports were compared with the expert reference standard. TP = true positive; TN = true negative; FP = false positive; FN = false negative.*

## Supplemental Table 9. Effect of AI Assistance on Diagnostic Performance for Secondary Findings

| Finding | Metric | Unassisted | Assisted | Difference percentage points |
|---|---|---|---|---|
| **Soft-tissue swelling 7 positive cases** | Accuracy | 324/348 (93.1%) | 331/348 (95.1%) | 2.0 |
| | Sensitivity | 3/21 (14.3%) | 14/21 (66.7%) | 52.4 |
| | Specificity | 321/327 (98.2%) | 317/327 (96.9%) | -1.2 |
| **Arthritis 14 positive cases** | Accuracy | 309/348 (88.8%) | 318/348 (91.4%) | 2.6 |
| | Sensitivity | 12/42 (28.6%) | 26/42 (61.9%) | 33.3 |
| | Specificity | 297/306 (97.1%) | 292/306 (95.4%) | -1.6 |
| **Prosthetic hardware 8 positive cases** | Accuracy | 336/348 (96.6%) | 340/348 (97.7%) | 1.1 |
| | Sensitivity | 20/24 (83.3%) | 22/24 (91.7%) | 8.3 |
| | Specificity | 316/324 (97.5%) | 318/324 (98.1%) | 0.6 |
| **Osteopenia 3 positive cases** | Accuracy | 333/348 (95.7%) | 327/348 (94.0%) | -1.7 |
| | Sensitivity | 0/9 (0.0%) | 9/9 (100.0%) | 100.0 |
| | Specificity | 333/339 (98.2%) | 318/339 (93.8%) | -4.4 |
| **Joint dislocation 3 positive cases** | Accuracy | 332/348 (95.4%) | 343/348 (98.6%) | 3.2 |
| | Sensitivity | 4/9 (44.4%) | 4/9 (44.4%) | 0.0 |
| | Specificity | 328/339 (96.8%) | 339/339 (100.0%) | 3.2 |

*Note.*—Values are pooled n/N (%) across three readers. All 116 examinations contributed to accuracy; sensitivity and specificity used reference-positive and reference-negative examinations, respectively. Differences are assisted minus unassisted. These exploratory outcomes are descriptive only: inferential confidence intervals, P values, and Holm-adjusted P values are not reported because of the small numbers of positive examinations. Osteopenia improved from 0/9 to 9/9 reader-case decisions, but these decisions arose from only three positive examinations and do not establish a precisely estimated population effect.

## Supplemental Table 10. Reader-Specific Performance for Fracture Detection With and Without Model Assistance

| Reader | Metric | Unassisted | Assisted | Difference percentage points |
|---|---|---|---|---|
| Reader 1 | Accuracy | 103/116 (88.8%) | 108/116 (93.1%) | 4.3 |
| Reader 1 | Sensitivity | 36/47 (76.6%) | 42/47 (89.4%) | 12.8 |
| Reader 1 | Specificity | 67/69 (97.1%) | 66/69 (95.7%) | -1.4 |
| Reader 2 | Accuracy | 100/116 (86.2%) | 109/116 (94.0%) | 7.8 |
| Reader 2 | Sensitivity | 34/47 (72.3%) | 45/47 (95.7%) | 23.4 |
| Reader 2 | Specificity | 66/69 (95.7%) | 64/69 (92.8%) | -2.9 |
| Reader 3 | Accuracy | 104/116 (89.7%) | 111/116 (95.7%) | 6.0 |
| Reader 3 | Sensitivity | 38/47 (80.9%) | 46/47 (97.9%) | 17.0 |
| Reader 3 | Specificity | 66/69 (95.7%) | 65/69 (94.2%) | -1.4 |

*Note.*—Values are n/N (%). Differences are assisted minus unassisted. Each reader evaluated 116 examinations, including 47 fracture-positive and 69 fracture-negative examinations. Reader-specific results are descriptive; no reader-specific hypothesis tests were performed.

## Supplemental Table 11. Comparison of Interpretation Time With and Without Model Assistance

### Interpretation time With and Without Assistance

| Readers | Condition | n | Median seconds | 25th percentile seconds | 75th percentile seconds |
|---|---|---|---|---|---|
| Pooled | Unassisted | 348 | 38.0 | 18.8 | 84.0 |
| Pooled | Assisted | 348 | 31.0 | 15.0 | 62.0 |
| Reader 1 | Unassisted | 116 | 62.0 | 29.0 | 139.5 |
| Reader 1 | Assisted | 116 | 52.5 | 27.3 | 108.5 |
| Reader 2 | Unassisted | 116 | 25.0 | 16.0 | 39.0 |
| Reader 2 | Assisted | 116 | 27.0 | 15.0 | 41.0 |
| Reader 3 | Unassisted | 116 | 39.0 | 13.0 | 72.0 |
| Reader 3 | Assisted | 116 | 19.5 | 11.0 | 55.5 |

## Adjusted comparison

| Effect | Time ratio | 95% confidence interval | Nominal P value |
|---|---|---|---|
| Assisted versus unassisted | 0.84 | (0.75, 0.94) | .004 |

*Note.*—The adjusted ratio is the assisted-to-unassisted geometric mean time ratio from ordinary least-squares regression of natural-log interpretation time on assistance, period, sequence, and categorical reader effects. Inference uses case-cluster-robust CR1 covariance and normal-Wald confidence limits and P values. All 348 reader–case pairs had two finite, positive times; no observations were excluded or truncated. No time-based outlier cutoff was applied. A ratio below 1 indicates shorter assisted interpretation time. The P value is nominal and exploratory. Reader-specific summaries are descriptive. The pooled analysis is conditional on the three readers and assumes independent examinations.

# Supplemental Methods.

## Supplemental Methods 1. Model Architecture

### Key architectural components of MSK-LLaVA.

| **Total parameters: 1.93B**<br>**LLM 1.72B, Image Encoder 200M, Connector 7.34M** | |
|---|---|
| **Component** | **Configuration** |
| **Visual encoder** | ConvNeXt-Large backbone from MSK-SigLIP. |
| **Visual encoder stages** | Four stages with depths [3, 3, 27, 3] and channel dimensions [192, 384, 768, 1536]. |
| **Multimodal projector** | Two-layer MLP: linear (1536 → 2048) → GELU → linear (2048 → 2048), with biases in both linear layers. |
| **Language model** | Qwen3-1.7B. |

## Supplemental Methods 2. Image Acquisition and Preprocessing

### Image acquisition

Musculoskeletal radiographs were retrospectively collected at a single institution from 2010 through 2023. The eligible internal source cohort comprised 1,404,105 radiographs. Equipment manufacturers were identified from DICOM header metadata and summarized by vendor. Philips systems accounted for 1,360,980 radiographs (96.9%); the remaining radiographs were acquired using equipment from other manufacturers or had unidentified or missing manufacturer information.

### Manufacturer distribution in the internal dataset

| Manufacturer | Number of radiographs | Percentage |
|---|---|---|
| Philips | 1,360,980 | 96.93 |
| AGFA | 23,796 | 1.69 |
| Canon | 9,879 | 0.70 |
| DRTECH | 4,035 | 0.29 |
| DK Medical Systems | 3,113 | 0.22 |
| Samsung Electronics | 466 | 0.03 |
| Carestream | 267 | 0.02 |
| FUJIFILM | 78 | 0.006 |
| Other, unidentified, or missing | 1,491 | 0.11 |
| **Total** | **1,404,105** | **100.00** |

Percentages use all 1,404,105 radiographs as the denominator. Values are rounded to two decimal places, except for FUJIFILM, which is shown to three decimal places because its contribution was less than 0.01%.

### DICOM intensity processing

DICOM files were read using pydicom, and stored pixel values were extracted using Dataset.pixel_array. Window center and window width were obtained from the WindowCenter and WindowWidth attributes, respectively. When multiple values were present, the first value of each attribute was selected. Pixel intensities were linearly mapped to the interval [0, 1] using lower and upper bounds equal to the window center minus and plus one-half of the window width, respectively. Values outside these bounds were clipped to 0 or 1. When either window attribute was absent, the conversion routine returned no image.

For radiographs with a PhotometricInterpretation value of MONOCHROME1, the normalized intensities were inverted by subtracting them from 1. No inversion was applied to MONOCHROME2 images. Windowing was applied directly to the stored pixel values, without a separate RescaleSlope/RescaleIntercept transformation or modality lookup-table operation.

### Spatial preprocessing and image storage

Each image was centered within a square array whose side length equaled the larger of the original image height and width. The remaining area was filled with zero-valued pixels, producing black padding while preserving the aspect ratio of the image content. The square image was resized to 1024 × 1024 pixels using bilinear interpolation (OpenCV INTER_LINEAR). Intensities were subsequently scaled to 0–65,535, clipped to this range, and stored as unsigned 16-bit grayscale images.

## Model input normalization

Images were loaded with MONAI (6) using PILReader and channel-first ordering, represented as floating-point arrays, and passed through a resize transform with a target size of 1024 × 1024 pixels. Intensity normalization used NormalizeIntensityd (channel_wise=False), with a subtrahend of 16,250.71 and a divisor of 17,877.68. These values corresponded to the pixel-intensity mean and standard deviation, respectively, estimated from 10,000 radiographs sampled from the training dataset. For the 8-bit input pathway, both parameters were divided by 256 relative to the 16-bit pathway. The normalized grayscale channel was replicated three times to provide a three-channel input to the image encoder. The preprocessing pipeline used pydicom version 2.4.4, NumPy version 1.26.3, OpenCV (opencv-python) version 4.9.0.80, Pillow version 10.3.0, and MONAI version 1.3.2.

## Training augmentation

Stochastic augmentation was applied only during training using MONAI transforms. Random zooming used factors between 0.8 and 1.1 with a probability of 1.0 and constant padding. Random in-plane rotation within ±60° was applied with a probability of 0.5 while retaining the image dimensions. Independent random flips along each spatial axis were applied with a probability of 0.5 per axis.

Gaussian noise was then applied using RandGaussianNoised with a probability of 0.5 and std=0.01. The transform order was zooming, rotation, optional flips, Gaussian noise, intensity normalization, and channel replication.

## External datasets and evaluation

External datasets were used in their distributed JPEG or PNG formats. After decoding, images were resized using interpolation and padding to match the input dimensions supported by each image encoder, and pixel values were normalized for model input. No stochastic augmentation was applied during validation, internal testing, or external evaluation.

# Supplemental Methods 3. Report Linkage and Anatomically Aligned Refinement

## Report linkage and refinement framework

Original radiologic reports corresponding to the radiographs were extracted from the institutional picture archiving and communication system (PACS). A report could describe multiple projections or anatomical regions, so its full text could contain findings unrelated to an individual image. We used a two-stage text-processing workflow consisting of examination-level standardization followed by image-level anatomical refinement. The resulting texts are referred to as anatomically aligned image-level reports.

## Model and inference settings

Report standardization and refinement used Llama 3.3 (70B) Instruct with Q8_0-quantized GGUF weights from the lmstudio-community distribution. Inference was performed using Ollama on two NVIDIA H100 graphics processing units. For examination-level generation, the configured temperature was 0.7, the nucleus sampling probability (top_p) was 0.9, and the repetition penalty (repeat_penalty) was 1.0.

## Examination-level standardization

The first-stage input comprised the examination protocol, the original abbreviated radiograph report, and the target body part. When an eligible CT or MRI report was available, its text and examination name were appended to the prompt as auxiliary context. The model was instructed to retain information relevant to the target body part, exclude findings not assessable on plain radiographs, and expand medical abbreviations. The prompt included two worked examples showing how to select relevant findings from reports that also described other anatomical regions.

## Selection and use of auxiliary CT and MRI reports

CT and MRI reports were eligible as auxiliary context only when the corresponding examination was performed on the same calendar date as the radiographic examination and matched the target body-part category. Anatomical matching was based on 15 categories: ankle, clavicle, elbow, femur, foot, forearm, hand, hip, upper arm, knee, lower leg, pelvis, scapula, shoulder, and wrist. The radiographic examination and the candidate CT or MRI examination were required to have the same assigned category.

Auxiliary CT or MRI report context was used in processing 21,806 radiograph–report pairs, corresponding to 1.55% of the 1,404,105 included radiographs. This proportion was calculated at the radiograph level.

The prompt instructed the model to exclude findings identifiable only on CT or MRI, including occult abnormalities not assessable on plain radiographs, and not to introduce such findings into the standardized radiograph report. These restrictions were implemented through prompt instructions. No dedicated quantitative audit was performed to assess the residual inclusion of CT- or MRI-only findings in the refined reports.

## Radiograph-level anatomical refinement

The second-stage input comprised the standardized clinical information, radiographic technique, findings, conclusion, and recommendation, together with the target image protocol and a protocol-specific list of anatomical structures. The protocol name conveyed the body part, laterality, and projection, such as “Ankle Right AP.” Structures were grouped as “Clearly visible structures” or “Partially visible or overlapping structures” to describe the expected anatomical coverage of the projection.

The prompt instructed the model to retain findings relevant to the target anatomy and laterality and to account for projection-dependent visibility. It also directed the model to exclude findings outside the target anatomical region and findings not assessable on plain radiographs. The model received text inputs only; no radiograph, CT, or MRI image pixels were provided at either stage. Accordingly, alignment reflected the report content and expected protocol coverage, and the procedure did not establish whether each finding was actually visible on an individual radiograph.

## Structured output and processing

Both prompts requested a JSON object containing report text, Boolean indicators, keywords, categorical finding labels, and interval change. Eight finding categories were assigned a requested vocabulary of “positive,” “negative,” “not mentioned,” or “uncertain.” Within this schema, “not mentioned” distinguished an unreported finding from an explicitly negative finding. Interval change was requested as “improved,” “worsened,” “no change,” “uncertain,” or “not mentioned.”

Outputs that could not be parsed as JSON were excluded from the processed report outputs. In the examination-level generation wrapper, if the parser returned more than one JSON object, only the first was retained. The schema describes the requested output format; successful JSON parsing alone does not establish compliance with all field and value requirements.

| JSON key | Type | Requested content or values |
|---|---|---|
| **clinical information** | String | Clinical context |
| **plain_radiography_technique** | String | Radiographic technique |
| **finding** | String | Findings for the target anatomy |
| **conclusion** | String | Summary or impression |
| **recommendation** | String | Recommendations |
| **further_study_needed** | Boolean | true or false |
| **is_fracture_exist** | Boolean | true or false |
| **abnormal** | Boolean | true or false |
| **keyword** | Array | Key terms |
| **fracture** | String | Four-state vocabulary* |
| **soft_tissue_swelling** | String | Four-state vocabulary* |
| **joint_effusion** | String | Four-state vocabulary* |
| **arthritis** | String | Four-state vocabulary* |
| **osteopenia** | String | Four-state vocabulary* |
| **foreign_body** | String | Four-state vocabulary* |
| **prosthetic_hardware** | String | Four-state vocabulary* |
| **joint_dislocation** | String | Four-state vocabulary* |
| **interval_change** | String | improved; worsened; no change; uncertain; not mentioned |

* Four-state vocabulary: “positive,” “negative,” “not mentioned,” or “uncertain.”

## Examination level standardization prompt

The templates below retain the original instructions and worked examples. Braced variable names denote runtime substitutions. Python f-string escape braces are displayed as literal JSON braces, and line wrapping is adjusted for readability.

The [ctmr_report_str] placeholder was empty when no CT or MRI report was supplied; otherwise, it contained the examination name followed by the report text.

```
<|begin_of_text|><|start_header_id|>system<|end_header_id|>
You are a skillful radiologist tasked with generating structured radiologic reports.
Write a standardized radiologic report for a plain radiograph based on the provided abbreviated X-ray report and CT/MRI report.
The output must be a valid JSON-formatted string with the following keys:
- "clinical information"
- "plain_radiography_technique"
- "finding"
- "conclusion"
- "recommendation"
- "further_study_needed" (boolean)
- "is_fracture_exist" (boolean)
```

- "abnormal" (boolean)
- "keyword" (list of key terms)
- "fracture"
- "soft_tissue_swelling"
- "joint_effusion"
- "arthritis"
- "osteopenia"
- "foreign_body"
- "prosthetic_hardware"
- "joint_dislocation"
- "interval_change" ("improved", "worsened", "no change", "uncertain", or "not mentioned")

For the following fields:
`fracture`, `soft_tissue_swelling`, `joint_effusion`, `arthritis`, `osteopenia`, `foreign_body`, `prosthetic_hardware`, `joint_dislocation`,
the value must be one of the following: `"positive"`, `"negative"`, `"not mentioned"`, or `"uncertain"`.

Instructions:
- Include only findings that are related to the target body part on plain radiograph.
- Do not include any findings not visible on plain radiograph (e.g., bone marrow edema).
- Ignore findings from other body parts.
- Assume that any relevant finding not mentioned is `"not mentioned"`.
- Include all critical information present in the input report.
- Detect whether the report mentions any interval change (e.g., improvement, worsening, or stability over time), and assign an appropriate value to `"interval_change"`.
- Avoid using medical abbreviations. All medical terms must be written in full.

Example Inputs and Outputs:
INPUT
Study protocol : Hip AP, Hip Lat
Target body part : hip
X-ray report : "Rt. proximal femur intertrochanteric fracture with displacement. There is joint effusion in right elbow. Equivocal Osteopenia."
Hip 3D CT report : "Rt. intertrochanteric fracture.
- Evans-Jensen classification type V, involving greater and lesser trochanter.
- externally rotated distal segment.

Clinical history: trauma
Comparison: no prior comparison available."

OUTPUT
```json
{
    "clinical information" : "trauma",
    "plain_radiography_technique" : "Anteroposterior and lateral plain radiographs of the hip were taken.",
    "finding" : "Right proximal femur shows an intertrochanteric fracture with displacement. The fracture involves the greater and lesser trochanter. The distal segment of the femur is externally rotated. ",
    "conclusion" : "The radiographic findings are consistent with a displaced right intertrochanteric fracture of the femur, involving both trochanters and associated with external rotation of the distal femoral segment. ",
    "recommendation" : "Orthopedic consultation for management of the right hip fracture. Follow-up imaging may be beneficial for monitoring fracture healing.",
    "further_study_needed" : true,
    "is_fracture_exist" : true,
    "abnormal" : true,
    "keyword" : ["fracture", "intertrochanteric fracture"],
    "soft_tissue_swelling" : "not mentioned",
    "joint_effusion" : "not mentioned",
    "arthritis" : "not mentioned",
    "osteopenia": "uncertain",
    "foreign_body": "not mentioned",
    "prosthetic_hardware": "not mentioned",
    "joint_dislocation": "not mentioned",
    "interval_change": "not mentioned"
}
```

INPUT
Study protocol : Ankle Rt. AP, Ankle Rt. Lat.
Target body part : ankle
X-ray report : "Right ankle shows a bimalleolar fracture with displacement. Lumbar spine shows spondylosis and old compression fracture of T12 vertebral body. Degenerative spondylolisthesis is noted at L4/5. Multilevel disc space narrowing is seen in the lumbar spine. Cortical irregularity is noted at the ulnar styloid process, but fracture cannot be ruled out. Chondrocalcinosis is seen in the right wrist. An accessory ossicle is noted distal to the ulnar styloid."

Clinical history: trauma
Comparison: no prior comparison available."

OUTPUT
```json
{
    "clinical information" : "trauma",
    "plain_radiography_technique" : "Anteroposterior and lateral plain radiographs of the right ankle were obtained.",
    "finding": "There is a displaced bimalleolar fracture in the right ankle.",
    "conclusion": "The radiographic findings are consistent with a displaced bimalleolar fracture of the right ankle.",
    "recommendation": "Orthopedic consultation for management of the right bimalleolar fracture. Follow-up imaging may be beneficial for monitoring fracture healing.",
    "further_study_needed" : true,
    "is_fracture_exist" : true,
    "abnormal" : true,
    "keyword" : ["fracture", "bimalleolar fracture"],
    "soft_tissue_swelling" : "not mentioned",
    "joint_effusion" : "not mentioned",
    "arthritis" : "not mentioned",
    "osteopenia": "not mentioned",
    "foreign_body": "not mentioned",
    "prosthetic_hardware": "not mentioned",
```

```
        "joint_dislocation": "not mentioned",
        "interval_change": "not mentioned"
    }

INPUT
    Study protocols : "[study_protocol]"
    Target body part : "[target_body_part]"
    X-ray report : "[abbreviated_report]"
    [ctmr_report_str]

    <|eot_id|><|start_header_id|>assistant<|end_header_id|>OUTPUT
    ```json
```

## Radiograph level refinement prompt

The [target_visible_structures] placeholder contains the protocol-specific anatomical list.

```
<|begin_of_text|><|start_header_id|>system<|end_header_id|>Be a helpful assistant.
You are a skillful radiologist tasked with generating structured radiologic reports.
You have a original radiologic report that covers multiple musculoskeletal X-rays.
Your task is to write a radiologic report for a single target X-ray based on the information provided in the original report
You must follow these principles:

Describe only the findings visible in the target X-ray.
If the original report describes a finding (e.g., fracture or swelling) in the right ankle, it implies that the finding is visible on the right ankle X-ray. If your target X-ray is for the left ankle, you should assume that the same finding is not visible, unless explicitly described for the left side.

Infer findings only if they are expected to be visible in the given X-ray view.
- For instance, a finding like "dorsal angulation" is not reliably seen in an anteroposterior (AP) view, so you must not describe it when interpreting an AP view.
Similarly:
- Joint space narrowing or osteophytes may be seen on weight-bearing or AP views, but not reliably on lateral views.
- Posterior malleolar fractures are more readily seen on lateral views than AP views.
- Alignment issues in the sagittal plane (e.g., dorsal or volar angulation) cannot be assessed from an AP view. Be mindful of the visibility of each finding in the actual projection view.

Make sure to describe any relevant findings for anatomical structures explicitly listed in the target protocol description.
- These are expected to be visible in the given projection.
- For example, if the protocol description states that the talus or ankle joint space is visible in the view, then those structures should be evaluated and included in the report.

Do not include findings outside of the target body part.
- For example, exclude tibial shaft abnormalities when reading an ankle-only view.

Do not describe findings that are not visible on plain radiographs.
- Findings such as bone marrow edema or cartilage defects should be excluded.

Assume any relevant but unmentioned findings as "not mentioned" in the structured output.
Check for and reflect any interval change (e.g., improvement, worsening, or no change) only if it is described in the report. Use "not mentioned" if no interval change is addressed.

The output must be a valid JSON-formatted string with the following keys:
- "clinical information"
- "plain_radiography_technique"
- "finding"
- "conclusion"
- "recommendation"
- "further_study_needed" (boolean)
- "is_fracture_exist" (boolean)
- "abnormal" (boolean)
- "keyword" (list of key terms)
- "fracture"
- "soft_tissue_swelling"
- "joint_effusion"
- "arthritis"
- "osteopenia"
- "foreign_body"
- "prosthetic_hardware"
- "joint_dislocation"
- "interval_change" ("improved", "worsened", "no change", "uncertain", or "not mentioned")

For the following fields:
`fracture`, `soft_tissue_swelling`, `joint_effusion`, `arthritis`, `osteopenia`, `foreign_body`, `prosthetic_hardware`, `joint_dislocation`,
the value must be one of the following: `"positive"`, `"negative"`, `"not mentioned"`, or `"uncertain"`.

Instructions:
- Include only findings that are related to the target body part on plain radiograph.
- Do not include any findings not visible on plain radiograph (e.g., bone marrow edema).
- Ignore findings from other body parts.
- Assume that any relevant finding not mentioned is `"not mentioned"`.
- Include all critical information present in the input report.
- Detect whether the report mentions any interval change (e.g., improvement, worsening, or stability over time), and assign an appropriate value to `"interval_change"`.
- Avoid using medical abbreviations. All medical terms must be written in full.

INPUT
Original X-ray clinical information : "trauma"
Original X-ray protocol : "Anteroposterior plain radiographs of the both ankles were taken."
Original X-ray report finding : "There is a displaced bimalleolar fracture in the right ankle. Osteopenia is noted on both ankles. Underlying severe osteoarthritis at both ankles."
Original X-ray report conclusion : "The radiographic findings are consistent with a displaced bimalleolar fracture of the right ankle."
```
```

```
Original X-ray report recommendation": "Orthopedic consultation for management of the right bimalleolar fracture. Follow-up imaging may be beneficial for monitoring fracture healing.",
Target X-ray protocol : "Ankle Lt. AP"
Target Protocol View Visible Structures :
{
    "Clearly visible structures": [
        "Distal tibial epiphysis",
        "Tibial plafond",
        "Medial malleolus",
        "Distal fibula",
        "Lateral malleolus",
        "Talus",
        "Talar dome",
        "Navicular bone",
        "Cuneiform bones (1st, 2nd, 3rd)",
        "Base of the 1st metatarsal",
        "Base of the 2nd metatarsal",
        "Base of the 3rd metatarsal",
        "Talocrural joint"
    ],
    "Partially visible or overlapping structures": [
        "Distal tibial diaphysis",
        "Distal fibular diaphysis",
        "Distal tibiofibular syndesmosis",
        "Calcaneus (superior aspect)",
        "Cuboid",
        "Base of the 4th metatarsal",
        "Base of the 5th metatarsal",
        "Posterior malleolus"
    ]
}

OUTPUT
```json
{
    "clinical information" : "trauma",
    "plain_radiography_technique" : "Anteroposterior plain radiograph of the left ankle was taken.",
    "finding" : "No acute fracture in the left ankle. Severe osteoarthritis at left ankle. Osteopenia is noted. ",
    "conclusion" : "Severe osteoarthritis at left ankle.",
    "recommendation" : "No further study is recommended.",
    "further_study_needed" : false,
    "is_fracture_exist" : false,
    "abnormal" : true,
    "keyword" : ["osteoarthritis", "osteopenia"],
    "soft_tissue_swelling" : "not mentioned",
    "joint_effusion" : "not mentioned",
    "arthritis" : "positive",
    "osteopenia": "positive",
    "foreign_body": "not mentioned",
    "prosthetic_hardware": "not mentioned",
    "joint_dislocation": "not mentioned",
    "interval_change": "not mentioned"
}
INPUT
Original X-ray clinical information : "[clinical_information]"
Original X-ray protocol : "[protocol]"
Original X-ray report finding : "[finding]"
Original X-ray report conclusion : "[conclusion]"
Original X-ray report recommendation": "[recommendation]",
Target X-ray protocol : "[target_protocol]"
Target Protocol View Visible Structures :
"[target_visible_structures]"

<|eot_id|><|start_header_id|>assistant<|end_header_id|>OUTPUT
```json
```

# Supplemental Methods 4. MSK-SigLIP Architecture and Training

## Overview

Model development comprised two stages for MSK-SigLIP image encoder training, followed by three stages for integration with a large language model and report-generation training. The image encoder was first adapted to musculoskeletal radiographs through supervised fracture classification and subsequently trained using image–text contrastive learning. The resulting encoder was incorporated into MSK-LLaVA through projector alignment, followed by full-model fine-tuning and additional fine-tuning on expert-annotated data. All training stages used bfloat16 (BF16) precision.

## MSK-SigLIP architecture

The visual backbone was ConvNeXt-Large, initialized with visual weights from the OpenCLIP checkpoint convnext_large_d_320 with the pretrained weight identifier laion2b_s29b_b131k_ft (Hugging Face repository: laion/CLIP-convnext_large_d_320.laion2B-s29B-b131K-ft). (7, 8) For image-text training, the visual encoder was paired with RadBERT (StanfordAIMI/RadBERT), a BERT-based text encoder pretrained on radiologic reports from a BioBERT initialization. (9) RadBERT comprises 12 transformer layers with a hidden dimension of 768. A text projection layer mapped the text representations into the shared image-text embedding space.

For both image encoder training stages, radiographs were prepared at a resolution of 1,024 × 1,024 pixels and stored as 16-bit grayscale images.

### Stage 1: Supervised fracture classification

The pretrained visual backbone was adapted to musculoskeletal radiographs by binary classification of fracture presence or absence using fracture labels for the full training dataset. Optimization used cross-entropy loss (torch.nn.CrossEntropyLoss) and stochastic gradient descent (SGD) with a learning rate of $1 \times 10^{-3}$. Training was performed for 55 epochs with a batch size of 16. This stage provided task-specific initialization for subsequent image-text training. Training was performed on four NVIDIA H100 GPUs and required 4 days and 39 minutes (96 hours and 39 minutes).

### Stage 2: Image-text contrastive training

The fracture-adapted visual encoder was subsequently trained within an image-text framework using the RadBERT text encoder and text projection layer. Each text input combined the imaging protocol with the corresponding anatomically aligned image-level report. These reports correspond to the image-specific refined reports described in the main text. Each image and its associated text constituted a positive pair. Image augmentation was applied during training, and the maximum text context length was 256 tokens.

Training used the sigmoid-based image-text objective implemented as SigLipLoss. (10) This objective encourages alignment of matched image-text pairs and separation of unmatched pairs through a pairwise sigmoid loss. Optimization used SGD with a base learning rate of $1 \times 10^{-2}$ and a global batch size of 512 for 34 epochs. The image and text encoders remained trainable throughout this stage, with no encoder freezing. A linear warm-up schedule (LinearLR) increased the learning rate from $10^{-6}$ times the base learning rate to the base learning rate over the first 800 training steps, followed by cosine annealing (CosineAnnealingLR) to a minimum learning rate of zero.

Model selection was based on top-3 image-to-text retrieval performance on the validation set of the expert-annotated dataset, and the best-performing checkpoint was retained as the final MSK-SigLIP model. Training was performed on four NVIDIA H100 GPUs and required approximately 299 hours.

### Stage 3A: Multimodal projector alignment

The trained MSK-SigLIP image encoder was connected to Qwen3-1.7B through a multimodal projection module to construct MSK-LLaVA. (11) This module mapped visual features to the language model embedding space. During the initial alignment stage, only the multimodal projector was optimized; the image encoder and language model remained frozen.

Projector alignment used the AdamW optimizer with a learning rate of $1 \times 10^{-3}$. Training was performed for one epoch with an effective global batch size of 256, a per-device mini-batch size of 4, and 64 gradient accumulation steps. The learning rate warm-up ratio was 0.03, and gradient checkpointing was disabled. Training used NVIDIA RTX PRO 6000 Blackwell Max-Q Workstation Edition hardware and required approximately 23 hours.

### Stage 3B: Full-model fine-tuning with single-image and multi-image inputs

Following projector alignment, MSK-LLaVA underwent full-model fine-tuning for report generation using both single-image and multi-image inputs. All model components, including the image encoder, multimodal projector, and language model, were jointly optimized using AdamW. The learning rate was $2 \times 10^{-6}$ for the image encoder and $1 \times 10^{-5}$ for the remaining model components. Training was performed for two epochs with a per-device mini-batch size of 1 and 128 gradient accumulation steps. Training used NVIDIA RTX PRO 6000 Blackwell Max-Q Workstation Edition hardware and required approximately 115 hours (4 days and 19 hours).

### Stage 3C: Additional fine-tuning on expert-annotated data

The model obtained after Stage 3B underwent an additional stage of full-model fine-tuning using the expert-annotated dataset with both single-image and multi-image inputs. All model components, including the image encoder, remained trainable. Optimization used AdamW with a learning rate of $2 \times 10^{-6}$ for the image encoder and $1 \times 10^{-5}$ for the remaining model components. Training was performed for one epoch with a per-device mini-batch size of 1 and 128 gradient accumulation steps. This stage used NVIDIA RTX PRO 6000 Blackwell Max-Q Workstation Edition hardware and required approximately 17 hours.

# Supplemental Methods 5. Encoder Evaluation

## Encoders and preprocessing

We evaluated frozen representations from MSK-SigLIP and seven public encoders by linear probing; the four vision-language models also underwent zero-shot classification. All models were evaluated without stochastic augmentation. Image features were converted to float32 and L2-normalized.

| Model | Checkpoint | Input size | Dim. | Representation |
|---|---|---|---|---|
| MSK-SigLIP | Local trained checkpoint | 1024 × 1024 | 768 | encode_image |
| MedSigLIP | google/medsiglip-448 (12) | 448 × 448 | 1152 | get_image_features |
| OpenAI CLIP | openai/clip-vit-large-patch14-336 (13) | 336 × 336 | 768 | Projected image embedding |
| SigLIP2 | google/siglip2-so400m-patch16-512 (14) | 512 × 512 | 1152 | get_image_features |
| RadioDINO | Snarcy/RadioDino-b16 (15) | 224 × 224 | 768 | timm forward output |
| SKELEX | skhoha/SKELEX (16) | 224 × 224 | 1024 | Final CLS token from model.vit |
| DINOv3 | facebook/dinov3-vit7b16-pretrain-lvd1689m (17) | 224 × 224 | 4096 | pooler_output |
| DINOv3-CNN | facebook/dinov3-convnext-large-pretrain-lvd1689m (17) | 224 × 224 | 1536 | pooler_output |

The first four encoders are vision-language models. Dim., feature dimension.

Images were loaded using MONAI PILReader and directly resized to 1024 × 1024 with area interpolation, with black padding. Intensity handling was configured as 16-bit for the internal dataset, MURA, and GRAZPEDWRI-DX, and 8-bit for LERA, FracAtlas, and BTXRD. MSK-SigLIP inputs were normalized as (x − 16,250.71)/17,877.68; both constants were divided by 256 for 8-bit inputs. Grayscale channels were replicated three times. For other encoders, intensities were clipped to the configured range, rescaled to unsigned 8-bit values, and replicated across three channels. MedSigLIP, CLIP, SigLIP2, and DINOv3 used checkpoint-associated processors. RadioDINO and SKELEX used 224 × 224 resizing and ImageNet normalization (means, 0.485/0.456/0.406; standard deviations, 0.229/0.224/0.225). SKELEX excluded the classification head; DINOv3 excluded patch and register-token concatenation.

## Datasets and evaluation units

The internal validation and expert-audited test files served as development and test sets, respectively. MURA used its supplied training and validation files. Other datasets were split approximately 90%/10% by patient (seed 42), keeping each patient's records together. Stratification used patient-level labels, positive if any record was positive, with single-label or multilabel stratification as appropriate; unstratified splitting was used when no variable label remained or a single-label class contained fewer than two patients.

| Dataset | Binary tasks | Evaluation unit |
|---|---|---|
| Internal | Fracture; arthritis; joint dislocation; osteopenia; prosthetic hardware; soft-tissue swelling | Radiograph |
| MURA; LERA | Abnormality | Examination |
| FracAtlas | Fracture | Radiograph |
| GRAZPEDWRI-DX | Fracture; osteopenia; cast; metallic hardware | Radiograph |
| BTXRD | Bone tumor present versus absent | Radiograph |
| BTXRD tumor subset | Malignancy | Radiograph |

For MURA and LERA, examination scores were arithmetic means of image scores, with identical reference labels required within each examination. Aggregation preceded regularization selection, threshold calibration, and final evaluation. Records without valid binary labels were excluded separately for each task.

## Zero-shot classification

Each class prototype was the L2-normalized mean of individually L2-normalized prompt embeddings. Cosine similarities to negative and positive prototypes ($s_-$ and $s_+$) yielded the positive-class score:

$$p_+ = \frac{exp(s_+)}{exp(s_-) + exp(s_+)}$$

No learned logit scale or bias was applied. These scores were not calibrated probabilities; scores ≥0.5 were classified as positive.

Internal tasks, MURA, LERA, and BTXRD prefixed each prompt with the exact protocol/body-part string and a newline; FracAtlas and GRAZPEDWRI-DX omitted this prefix. Internal tasks and GRAZPEDWRI-DX/BTXRD tumor detection used three prompts per class:

Negative: “No [x] is present.”; “No [x] is identified.”; “There is no [x].”

Positive: “[Y] is present.”; “[Y] is identified.”; “There is [y].”

Here, y is “a [x]” for fracture, joint dislocation, cast, and bone tumor, and otherwise equals x; Y capitalizes the first letter of y. Task-specific substitutions were:

| Dataset | Task | [x] |
|---|---|---|
| Internal; GRAZPEDWRI-DX | Fracture | fracture |
| Internal | Soft-tissue swelling | soft tissue swelling |
| Internal | Arthritis | arthritis |
| Internal | Joint dislocation | joint dislocation |
| Internal | Osteopenia | osteopenia |
| Internal | Prosthetic hardware | orthopedic hardware or prosthesis |
| GRAZPEDWRI-DX | Osteopenia; cast; metallic hardware | osteopenia; cast; metallic hardware, respectively |
| BTXRD | Tumor | bone tumor |
| BTXRD tumor subset | Malignancy | malignant bone tumor |

FracAtlas used the single negative/positive prompts “No fracture”/“Fracture”. MURA and LERA used “no abnormal finding”/“abnormal finding (fracture, soft tissue swelling, prosthetic hardware, Osteopenia, joint dislocation)”. BTXRD malignancy prompts were “The bone tumor is benign.”, “A benign bone tumor is present.”, and “This is a benign bone tumor.”; positive prompts replaced “benign” with “malignant”.

## Linear probing

Each task used L2-regularized scikit-learn LogisticRegression (lbfgs; no class weighting; maximum 2000 iterations; tolerance $10^{-12}$; random_state=42), without feature standardization or dimensionality reduction. All fitting, regularization selection, and threshold calibration used development data only.

The inverse regularization parameter C was selected from 45 logarithmically spaced values from $10^{-6}$ to $10^{5}$ using stratified grouped cross-validation (shuffle=True; random_state=42), preferentially grouped by patient. Five folds were requested and reduced to four, three, or two if any training or validation fold lacked either class; tasks without two valid folds were not evaluated. C maximized AUC of pooled out-of-fold predictions, after examination aggregation where applicable, with ties favoring smaller C. The corresponding pooled predictions determined the F1-maximizing threshold; ties used the first threshold returned by the precision-recall routine. The classifier was refitted on all eligible development data and evaluated on the test set with this threshold fixed. Out-of-fold estimates were not considered independent test results.

## Statistical analysis

We calculated task-specific AUC and F1 using the method-specific threshold. Sensitivity, specificity, predictive values, accuracy, and balanced accuracy were also computed. Percentile 95% confidence intervals used 2000 bootstrap resamples of evaluation records, stratified by class when both classes were present. Resampling units were examinations for MURA/LERA and radiographs otherwise, without patient-cluster resampling. Classifiers and thresholds remained fixed. NumPy default_rng used seed 42 plus the zero-based task index. Nonfinite estimates were excluded per metric; AUC was undefined for single-class samples. Test estimates were reported separately from development estimates.

# Supplemental Methods 6. Comparator Model Evaluation

Seven publicly available vision–language models were evaluated without additional fine-tuning for musculoskeletal report generation. Model identifiers and model-specific inference settings are summarized below.

All models received the same radiographs and semantically equivalent instructions, using model-specific image preprocessing and chat templates. Examinations were processed individually in a fixed order. MSK-LLaVA received the collated image tensors directly. Three comparators were loaded directly from Hugging Face checkpoints, and the remaining four were evaluated through Ollama. Although the Gemma 3 architecture used by MedGemma supports multiple input radiographs, the MedGemma 27B model card states that multi-image comprehension had not been evaluated.

MSK-LLaVA and the three Hugging Face-based comparators used greedy decoding with do_sample=False. For MSK-LLaVA, top_p=0.9 and top_k=0 were configured but inactive because sampling was disabled. For the Ollama-based comparators, the wrapper requested temperature=0.0, and top_p=0.9.

| Model | Checkpoint or inference identifier | Parameters | Image resolution / preprocessing | Precision | Reasoning configuration | Repetition penalty |
|---|---|---|---|---|---|---|
| **MSK-LLaVA** | Local checkpoint | 1.9B | 1024 × 1024 | BF16 | Off | 1.0 |
| **Gemma 4 31B IT** | gemma4:31b-it-bf16 | 31B | 1024 × 1024 | BF16 | On | 1.0 |
| **MedGemma** | medgemma:27b-it-bf16 | 27B | 896 × 896 | BF16 | No separate | 1.0 |
| **27B IT** | | | | | mode | |
| **MedGemma 1.5 4B IT** | medgemma1.5:4b-it-bf16 | 4B | 896 × 896 | BF16 | No separate mode | 1.05 |
| **Qwen3.8 27B** | qwen3.8:27b-bf16 | 27B | 1024 × 1024 | BF16 | On | 1.0 |
| **Lingshu-32B** | lingshu-medical-mllm/Lingshu-32B | 32B | 1036 × 1036, resized from 1024 × 1024 | BF16 | No separate mode | 1.05 |
| **Fleming-VL-38B** | IQuestLab/Fleming-VL-38B | 38B | 448 × 448; nine tiles and one thumbnail per source image | BF16 | Native template; no additional CoT prompt | 1.0 |
| **MedDr-40B** | Sunanhe/MedDr_0401 | 40B | 448 × 448 | BF16 | No separate mode | 1.05 |

# Supplemental Methods 7. Structured Report Labeler

## Purpose and extraction workflow

GPT-4o mini (OpenAI; gpt-4o-mini-2024-07-18; temperature, 0; top-p, 1.0) extracted six findings from generated reports: fracture, soft-tissue swelling, arthritis, prosthetic hardware, osteopenia, and joint dislocation. Only generated report text was provided, without radiographs or reference reports. The prompt requested JSON labels of positive (explicitly present), negative (explicitly absent), uncertain (possible or suspected), or missing (unmentioned). For binary evaluation, positive and uncertain labels were mapped to 1, and negative and missing labels to 0. The prompt instructed the labeler to assign negative labels to all six findings for “no significant bony abnormality” or equivalent statements and to extract bone, laterality, and anatomical location for fractures. Labeling identified findings expressed in the text without assessing overall report quality or concordance with radiographs.

## Human validation and agreement

Two musculoskeletal radiologists independently validated the automatically extracted finding labels for 100 radiologic reports generated by MSK-LLaVA for examinations in the internal test set. Each radiologist reviewed the report text together with its automatically extracted labels. Disagreements between the radiologists were resolved by consensus. Agreement between the automated labels and the consensus human assessments was evaluated at the report-finding level after binary conversion: positive and uncertain were mapped to positive, whereas negative and missing were mapped to negative. Six findings per report were evaluated, yielding 600 binary comparisons.

Automated labels agreed with the consensus human assessments in 598 of 600 comparisons (99.67%). Both discordant labels involved reports describing a foreign body, which the automated labeler incorrectly classified as positive for prosthetic hardware. Agreement was 98/100 (98%) for prosthetic hardware and 100/100 (100%) for each of the other five findings. The original automated labels, including the two discordant labels, were retained without manual correction in subsequent performance analyses. This validation assessed the extraction of findings from report text; it did not establish the diagnostic accuracy or clinical quality of the generated reports.

| Finding | Reviewed labels | Discordant labels | Agreement |
|---|---|---|---|
| Fracture | 100 | 0 | 100.0% |
| Soft-tissue swelling | 100 | 0 | 100.0% |
| Arthritis | 100 | 0 | 100.0% |
| Prosthetic hardware | 100 | 2 | 98.0% |
| Osteopenia | 100 | 0 | 100.0% |
| Joint dislocation | 100 | 0 | 100.0% |
| Overall | 600 | 2 | 99.7% |

# Supplemental Methods 8. Grad-CAM and Reader Interface

For reader assistance, fracture heatmaps were generated from the stage 1 fracture classifier used to initialize the MSK-SigLIP image encoder. Each radiograph was processed individually at 1024 × 1024 pixels. Grad-CAM (18) targeted the fracture-positive class score at the final encoder block. Heatmaps were resized and overlaid using the JET colormap, with opacity determined by the predicted fracture probability and set to zero for probabilities <0.50. A web-based interface displayed generated reports and heatmap-overlaid radiographs, allowing readers to toggle overlays and adjust zoom and window settings.

For illustrative figures, separate Grad-CAM maps were generated from the final MSK-LLaVA model at the last stage of its image encoder, with all examination radiographs supplied jointly. Targets were the mean token log probability of selected generated text, preserving the preceding report context. Fracture and osteoarthritis maps used the difference from the corresponding negative alternatives, “no fracture” and “no arthritis,” respectively; the calcific tendinitis map used the selected statement alone. Maps were rectified, normalized using a shared scale across views, and overlaid on the radiographs. These maps were not used in the reader study.

# Supplemental Methods 9. Multireader Randomization and Statistical Analysis

## Study design and analysis population

Three nonradiologist physicians each interpreted 116 external examinations (47 fracture-positive and 69 fracture-negative) with and without MSK-LLaVA. A shared, nonstratified case-level schedule (seed 42) allocated 58 examinations to each sequence. Case order was randomized within sessions separated by 3 weeks. Readers had access to age and sex but not previous responses or reference-standard reports. The analysis contained 348 complete reader–case pairs and 696 interpretations; no selected finding or interpretation-time record was excluded.

The primary endpoint was the absolute difference in fracture accuracy. Fracture sensitivity, specificity, and interpretation time were supporting outcomes. The five other manuscript findings were summarized descriptively. No a priori power calculation was performed.

## Paired performance comparisons

For each examination, correctness was averaged across the three readers separately under each condition. The assisted-minus-unassisted difference was then calculated. The overall effect was the mean of these examination-level differences; its standard error was their sample standard deviation divided by the square root of the number of examinations. Two-sided t tests and 95% t intervals used 115 degrees of freedom for accuracy, 46 for sensitivity, and 68 for specificity. Sensitivity and specificity used the 47 reference-positive and 69 reference-negative examinations, respectively.

Descriptive pooled-proportion CIs used 20,000 percentile bootstrap replicates, resampling examinations with replacement while retaining all three readers and both conditions. The generator was reinitialized with seed 20260411 for each endpoint. Bootstrap intervals for the differences were also calculated as a robustness check. Readers were not resampled; inference was conditional on the participating readers and assumed independent examinations. Nonfracture findings were reported as counts, proportions, and absolute differences, without inferential intervals or P values because positive-case counts were small.

## Adjusted analyses and software

Separate logistic generalized estimating equations for fracture accuracy, sensitivity, and specificity included assistance, numerical period, sequence, and categorical reader effects. Working independence and the empirical case-cluster sandwich covariance were used. Odds ratios and 95% normal-Wald CIs were obtained by exponentiating the assistance coefficient and its limits. Rank and convergence were checked. Neither a separate carryover term nor an assistance-by-period interaction was fitted.

Interpretation time was reduced to one record per reader–case–condition; only pairs with two finite, positive times were eligible. Times were natural-log transformed without truncation. Ordinary least-squares regression used the same covariates and case-cluster CR1 covariance, with normal-Wald inference. Exponentiated coefficients yielded geometric mean time ratios. Descriptive times were medians and interquartile ranges.

The primary accuracy test used two-sided $P < .05$. Supporting P values were nominal and exploratory; no Holm adjustment was applied because nonfracture findings were descriptive. Computations used Python 3.12.14, NumPy 2.3.5, pandas 2.2.3, and SciPy 1.17.0. Logistic score equations and cluster covariance estimates were implemented explicitly and checked numerically.